\documentclass[conference]{IEEEtran}
\usepackage{times}

\usepackage[numbers]{natbib}
\usepackage{multicol}
\usepackage{graphicx}
\usepackage{float}
\usepackage{amsmath}
\usepackage{amssymb}
\usepackage{array}
\usepackage{booktabs}
\usepackage{multicol,multirow}
\usepackage{setspace}
\usepackage{makecell}
\usepackage{fontawesome}
\usepackage[dvipsnames]{xcolor}
\usepackage{tasks}
\usepackage{pifont}
\usepackage{colortbl}
\usepackage{xcolor}
\usepackage{framed}
\usepackage{wrapfig}  
\usepackage{titletoc}
\usepackage{color} 
\usepackage{caption, subcaption, overpic, textpos}
\usepackage{titlesec}
\usepackage{comment}

\makeatletter
\DeclareRobustCommand\onedot{\futurelet\@let@token\@onedot}
\def\@onedot{\ifx\@let@token.\else.\null\fi\xspace}

\def\eg{\emph{e.g}\onedot}

\makeatother

\usepackage{xspace}
\newcommand{\modelname}{UniVLA\xspace}

\definecolor{deemph}{gray}{0.6}

\definecolor{baselinecolor}{gray}{.9}
\definecolor{yellow}{RGB}{218,165,32}
\definecolor{lightcyan}{rgb}{0.88, 1.0, 1.0}
\definecolor{lightskyblue}{rgb}{0.53, 0.81, 0.98}
\definecolor{aliceblue}{rgb}{0.94, 0.97, 1.0}
\definecolor{LightSlateBlue}{RGB}{70,130,180}
\definecolor{DeepBlue}{RGB}{65,100,170}
\definecolor{DeepPurple}{RGB}{136,105,160}
\definecolor{LightGreen}{RGB}{59,125,35}
\definecolor{LightRed}{RGB}{234,66,53}
\definecolor{cvprblue}{rgb}{0.21,0.49,0.74}
\newcommand{\baseline}[1]{\cellcolor{aliceblue}{#1}}
\usepackage[pagebackref,breaklinks,colorlinks,citecolor=cvprblue,bookmarks=true]{hyperref}

\usepackage[capitalize]{cleveref}

\crefname{section}{Sec.}{Secs.}
\crefname{section}{Sec.}{Secs.}

\crefname{figure}{Fig.}{Figs.} %
\Crefname{figure}{Fig.}{Figs.} %

\crefname{table}{Tab.}{Tabs.} %
\Crefname{table}{Tab.}{Tabs.} %

\usepackage{epigraph}
\newcommand{\underfigtab}{\vspace{-10pt}}

\newlength\savewidth

\renewcommand{\paragraph}[1]{\vspace{1.25mm}\noindent\textbf{#1}}

\newcolumntype{x}[1]{>{\centering\arraybackslash}p{#1pt}}
\newcolumntype{y}[1]{>{\raggedright\arraybackslash}p{#1pt}}
\newcolumntype{z}[1]{>{\raggedleft\arraybackslash}p{#1pt}}

\newcommand{\app}{\raise.17ex\hbox{$\scriptstyle\sim$}}

\begin{document}

\title{Learning to Act Anywhere with \\ Task-centric Latent Actions}

\author{\authorblockN{
Qingwen Bu$^{1,2}$,
Yanting Yang$^{2}$,
Jisong Cai$^{2}$,
Shenyuan Gao$^{2}$, 
Guanghui Ren$^{3}$, \\
Maoqing Yao$^{3}$,
Ping Luo$^{1,2}$
and
Hongyang Li$^{1,2}$
}
\smallskip
\authorblockA{
$^{1}$ The University of Hong Kong
$^{2}$ OpenDriveLab
~$^{3}$ AgiBot \\
{\small \faGithub\ Code: \texttt{\url{https://github.com/OpenDriveLab/UniVLA}}}}
}

\noindent
\twocolumn[{%
\renewcommand\twocolumn[1][]{#1}
\maketitle
\vspace{-5mm}
\begin{center}
    \centering
    \captionsetup{type=figure}
    \includegraphics[width=0.99\textwidth]{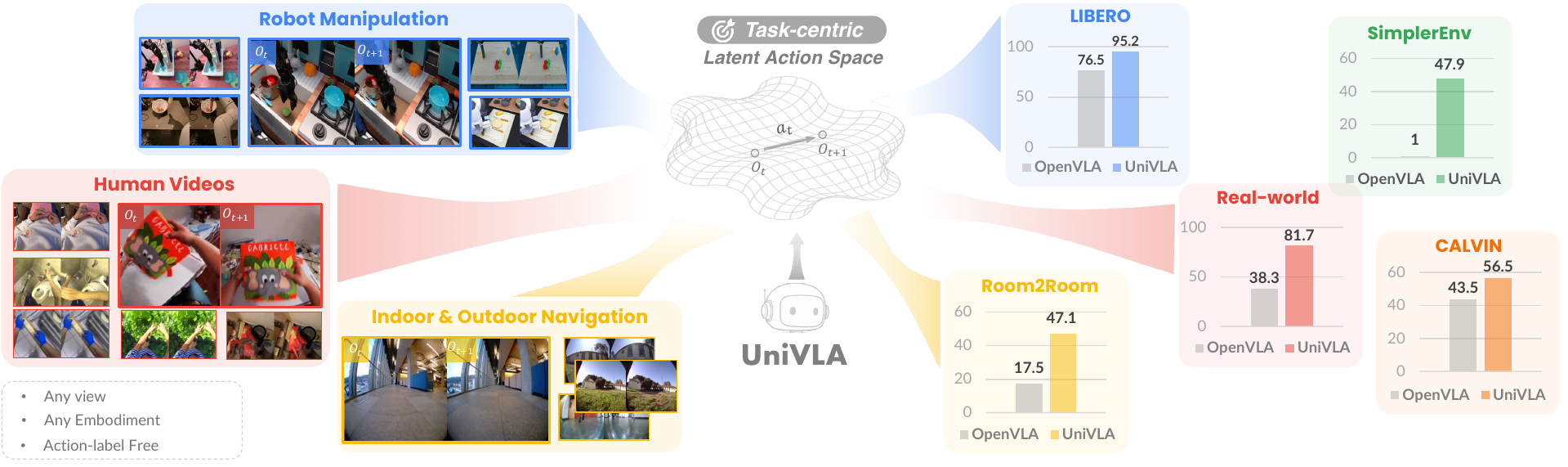}
    \captionof{figure}{
    We introduce \textbf{\modelname}, a unified vision-language-action (VLA) framework that enables policy learning across different environments. By deriving task-centric latent actions in an unsupervised manner, \modelname can leverage data from arbitrary embodiments and perspectives without action labels. After large-scale pretraining from videos, \modelname develops a cross-embodiment generalist policy that can be readily deployed across various robots by learning an action decoding with minimal cost. Compared to OpenVLA~\cite{kim2024openvla}, \modelname exhibits unanimous improvement on multiple manipulation and navigation tasks.
    \label{fig:teaser}
    \underfigtab
    }
\end{center}%
}]

\begin{abstract}
A generalist robot should perform effectively across various environments. 
However, most existing approaches heavily rely on scaling action-annotated 
data
to enhance their capabilities. Consequently, they are often limited to single physical specification and struggle to learn transferable knowledge across different embodiments and environments. 
To confront these limitations, we propose \modelname, a new framework for learning cross-embodiment vision-language-action (VLA) policies. 
Our key innovation is to derive task-centric action representations from videos with a latent action model. This enables us to exploit extensive data across a wide spectrum of embodiments and perspectives. 
To mitigate the effect of task-irrelevant dynamics, we incorporate language instructions and establish a latent action model within the DINO feature space. 
Learned from internet-scale videos, the generalist policy can be deployed to various robots through efficient latent action decoding.
We obtain state-of-the-art results across multiple manipulation and navigation benchmarks, as well as real-robot deployments.
\modelname achieves superior performance over OpenVLA with less than 1/20
of pretraining compute and 1/10 of downstream data. Continuous performance improvements are observed as heterogeneous data, even including human videos, are incorporated into the training pipeline. The results underscore UniVLA's potential to facilitate scalable and efficient robot policy learning.

\end{abstract}

\IEEEpeerreviewmaketitle

\section{Introduction}

Empowered by the emergence of large-scale robotic datasets~\cite{walke2023bridgedata,padalkar2023open,khazatsky2024droid,contributors2025agibotworld}, robot policies based on vision-language-action models (VLA) have made encouraging strides recently~\cite{brohan2023rt2,team2024octo,kim2024openvla}. However, they typically rely on ground-truth action labels for supervision, which limits their scalability in utilizing internet-scale data from diverse environments. Furthermore, the heterogeneity of action and observation spaces across different embodiments (\eg, Franka, WidowX, and even human hands) and tasks (\eg, manipulation and navigation) poses a significant challenge to effective knowledge transfer. 
This raises a crucial question: could we learn a \textit{unified action representation} that enables the generalist policy to plan effectively, unlocking the potential of internet-scale videos and facilitating knowledge transfer \textit{across different embodiments and environments}?

To address these challenges, we propose \modelname, a generalist policy learning framework that enables scalable and efficient planning across various embodiments and environments. 
Much like large language models (LLMs) learn cross-lingual shared knowledge~\cite{devlin2018bert,conneau2019unsupervised}, we aim to construct a unified action space that facilitates knowledge transfer across video data, including various robot demonstrations and egocentric human videos.
Our recipe for generalist policy consists of three key stages:
\textbf{1) Task-centric Latent Action Learning}, where we extract task-relevant action representations from massive cross-embodiment videos in an unsupervised manner. This is achieved by discretizing latent actions from the inverse dynamics of paired frames using a VQ-VAE~\cite{van2017neural}. %
\textbf{2) Next-latent action prediction}, where we train an auto-regressive vision-language model with discretized latent action tokens, endowing it with embodiment-agnostic planning capabilities. 
\textbf{3) Latents decoding}, where we decode latent plans into physical behaviors and specialize the pretrained generalist policy for deployment in unseen tasks efficiently.

While recent studies~\cite{ye2024lapa,chen2024igor} have investigated the viability of learning latent actions from web-scale videos, they suffer from a critical limitation: their naive reconstruction-based objectives often capture task-irrelevant dynamics, such as movements of non-ego agents or unpredictable camera shifts. These noisy representations hinder policy pretraining by introducing distractions unrelated to the task. To address this, we leverage pre-trained DINOv2 features~\cite{oquab2023dinov2} to extract patch-level representations from pixels, providing both spatial and object-centric priors that better capture task-relevant information. By using the readily available language instructions as conditions, we further disentangle movements into two complementary action representations, one of which explicitly represents task-centric actions.

\modelname achieves state-of-the-art performance across multiple manipulation benchmarks and navigation tasks, outperforming OpenVLA~\cite{kim2024openvla} by a significant margin while requiring merely 1/20 of the pretraining cost (in GPU hours). This efficiency stems from its task-centric latent action space, which decouples task-relevant dynamics from extraneous visual changes. 
Our action representation not only reduces computational overhead but also enables efficient scaling - as dataset size grows, \modelname's performance improves, effectively leveraging cross-embodiment, cross-view robot datasets and even unlabeled human videos to expand its pretraining corpus and extract transferable knowledge.
Remarkably, when pretrained solely on the Bridge-V2 dataset~\cite{walke2023bridgedata}, \modelname surpasses OpenVLA and LAPA 
trained on the larger Open X-Embodiment~\cite{padalkar2023open} dataset, underscoring its ability to distill transferrable knowledge from \textit{limited} data.

In addition, we employ a lightweight decoder with only 10.8M parameters to translate latent actions into executable trajectories, significantly reducing the need for extensive fine-tuning. This design leverages the compact and informative nature of the task-centric latent action space, enabling \modelname to adapt efficiently to diverse tasks and embodiments with minimal downstream data. Our comprehensive evaluation, spanning manipulation, navigation, and real-world deployment, underscores the framework’s efficiency, scalability, and generalizability, positioning it as a promising pathway toward next-generation generalist robotic policies. 

In summary, our main \textbf{contributions} are three-folds:

\begin{itemize}
    \item We propose \modelname, a recipe towards generalist policy by planning in a unified, embodiment-agnostic action space, enabling scalable and efficient decision-making by learning from web-scale videos.

    \item  We introduce a novel approach for extracting task-relevant latent actions from cross-embodiment videos, decoupling task-centric dynamics from irrelevant visual changes. Both qualitative and quantitative experiments highlight its merits and advantages over existing works.

    \item \modelname achieves state-of-the-art performance on multiple benchmarks and real-robot tests, achieving an 18.5\% increase in success rate over OpenVLA on the LIBERO~\cite{liu2024libero} benchmark, 29.6\% in navigation tasks~\cite{anderson2018r2r}, and a 36.7\% improvement in real-world deployments.
\end{itemize}

\section{Related Work}

\subsection{Vision-language-action Models}

Building on the success of pretrained vision foundation models, large language models (LLMs), and vision-language models (VLMs), VLAs have been introduced to process multimodal inputs—visual observations and language instructions—and generate robotic actions for completing embodied tasks. RT-1~\cite{brohan2022rt1} and Octo~\cite{team2024octo} employ a transformer-based policy that integrates diverse data, including robot trajectories across various tasks, objects, environments, and embodiments. In contrast, some prior works~\cite{brohan2023rt2,kim2024openvla,li2023roboflamingo} leverage pretrained VLMs to generate robotic actions by tapping into world knowledge from large-scale vision-language datasets. For instance, RT-2~\cite{brohan2023rt2} and OpenVLA~\cite{kim2024openvla} treat actions as tokens within the language model’s vocabulary, while RoboFlamingo~\cite{li2023roboflamingo} introduces an additional policy head for action prediction. Building on these generalist policies, RoboDual~\cite{bu2024robodual} proposes a synergistic dual-system that combines the strengths of both generalist and specialist policy. Other works incorporate goal image~\cite{black202susie} or video~\cite{du2024UniPi,wu2023gr1,bu2024clover} prediction tasks to generate valid, executable plans conditioned on language instructions, with these visual cues subsequently guiding the policy in action generation.
However, these methods heavily rely on interactive data with ground-truth action labels, which significantly restricts the scalability of VLAs. In contrast, our approach unlocks the potential of internet-scale, action-free videos by learning a unified latent action representation from visual changes, independent of action labels.

\subsection{Cross-embodiment Learning}
Training a general-purpose robot policy is challenging due to the diversity in camera perspectives, proprioceptive inputs, joint configurations, action spaces, and control frequencies across robotic systems. Early approach~\cite{yang2024pushing} focused on aligning action space manually between navigation and manipulation but were limited to wrist cameras in manipulation. Recent transformer-based approaches~\cite{team2024octo, Doshi24-crossformer} address these challenges by accommodating variable observations and actions, with CrossFormer~\cite{Doshi24-crossformer} co-training across four distinct action spaces without imposing constraints on observation spaces or requiring explicit action-space alignment. Flow representations, capturing future trajectories of query points in images or point clouds, have been widely explored for cross-embodiment learning~\cite{wen2023any,yuan2024general,gao2024flip,xu2024flow}. ATM~\cite{wen2023any} learns flow generation from human demonstrations, while Im2Flow2Act~\cite{xu2024flow} predicts object flows from human videos without in-domain data. Meanwhile, object-centric representations~\cite{hsu2024spot, bharadhwaj2024hopman} offer an alternative approach, with SPOT~\cite{hsu2024spot} predicting object trajectories in SE(3) to decouple embodiment actions from sensory inputs.
Existing approaches demand extensive, diverse datasets to cover all possible state-transition patterns and need explicit annotations, leading to inefficient data utilization. Our method sets itself apart by using a discrete codebook to encode latent actions in an unsupervised manner. Our approach effectively filters out visual noise and achieves efficient information compression via vector quantization, thereby enhancing training efficiency and lessening the reliance on data diversity.

\subsection{Latent Action Learning}
Several prior works focus on learning variational auto-encoders~\cite{pu2016variational, van2017neural} on raw action trajectories to structure new action spaces, emphasizing compact latent representations that facilitate behavior generation and task adaptation, as seen in VQ-BeT~\cite{lee2024behavior} and Quest~\cite{mete2024quest}. These methods are also adopted in reinforcement learning to accelerate convergence~\cite{allshire2021laser}. Recent works~\cite{wang2024omnijarvis,szot2024grounding} explore vector quantization as action space adapters to better integrate actions into large language models. However, a key limitation of these approaches is their reliance on ground-truth action labels, which limits their scalability.

To leverage broader video data, Genie~\cite{bruce2024genie} extracts latent actions via a causal latent action model, conditioning on next-frame prediction. Similarly, LAPO~\cite{schmidt2024lapo} and DynaMo~\cite{cui2024dynamo} learn latent actions directly from visual data, bypassing methods using explicit action labels on in-domain manipulation tasks. LAPA~\cite{ye2024lapa} and IGOR~\cite{chen2024igor} introduce unsupervised pretraining methods to teach VLAs discrete latent actions, aiming to transfer knowledge from human videos. However, these approaches encode all visual changes from raw pixels, capturing task-irrelevant dynamics such as camera shakiness, movements of other agents, or new object appearances, which ultimately degrade policy performance. We propose a novel training framework to decouple task-centric dynamics from irrelevant visual changes, structuring a more effective latent action space to enable robust policy planning.

\begin{figure*}[t]
    \centering
    \includegraphics[width=0.97\linewidth]{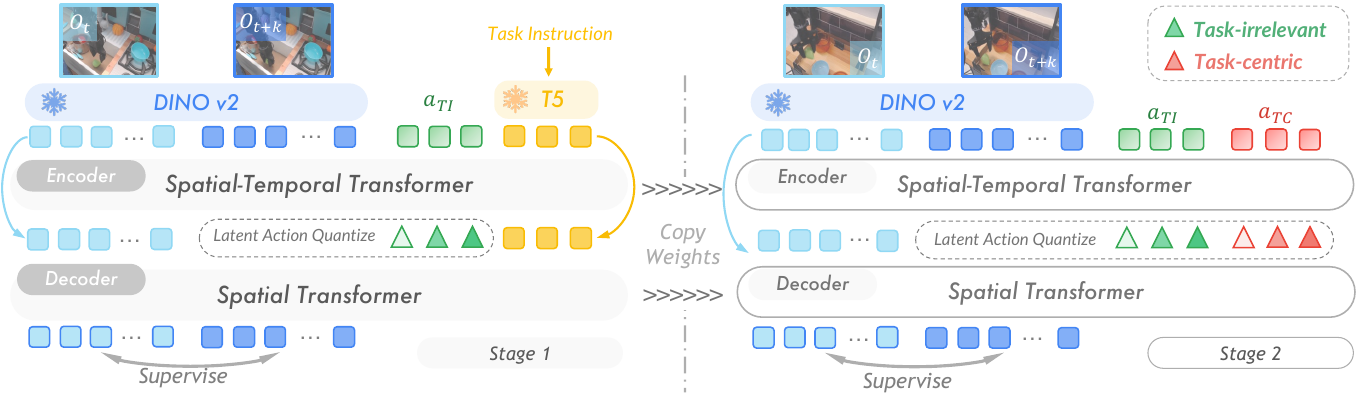}
    \caption{\textbf{Task-centric latent action learning.} We propose a two-stage training framework aimed at disentangling task-centric visual dynamics and changes from extraneous factors. In \textbf{Stage 1}, task instruction embeddings, derived from a pre-trained T5 text encoder~\cite{raffel2020exploring}, are utilized as inputs to both the encoder and decoder. These embeddings provide task-relevant semantic information to enhance predictive accuracy. In \textbf{Stage 2}, a novel set of latent actions is introduced, specifically designed to replace the role of language and to capture task-centric dynamics from DINOv2-encoded features of video frames.}
    \vspace{-7pt}
    \label{fig:lam}
\end{figure*}

\medskip
\section{Methodology}

We develop three steps to implement \modelname: \textbf{1)} (\Cref{sec:step1}) Leveraging language-based goal specifications, we extract inverse dynamics from extensive video datasets in an unsupervised manner, yielding a discretized set of task-centric latent actions that generalize across diverse embodiments and domains; \textbf{2)} (\Cref{sec:step2}) Based on this, we train an auto-regressive transformer-based vision-language-action model, which takes visual observations and task instructions as inputs to predict latent action tokens in a unified latent space; \textbf{3)} (\Cref{sec:step3}) To facilitate efficient adaptation to various robotic control systems, we introduce specialized policy heads that decode latent actions into executable control signals.

\subsection{Task-centric Latent Action Learning}
\label{sec:step1}

The first step establishes the foundational groundwork of our framework by generating the pseudo action labels (\textit{i.e.}, latent action tokens), which serve as the basis for training our generalist policy in subsequent stages.

\paragraph{Latent action quantization.} \Cref{fig:lam} illustrates the two-stage training pipeline and overall architecture of our latent action model. We start with a pair of consecutive video frames, denoted as $\{o_{t}, o_{t+k} \}$, separated by a frame interval $k$. 
To ensure a uniform time interval of approximately 1 second across diverse datasets, the frame interval is calibrated according to the recording frequency specific to each dataset. To derive latent actions from videos, our latent action model is constructed around an Inverse Dynamics Model (IDM) based encoder $\mathcal{I}(a_{t} | o_{t}, o_{t+k})$ and a Forward Dynamics Model (FDM) based decoder $\mathcal{F}(o_{t+k} | o_{t}, a_{t})$. The encoder infers latent action given consecutive observations, and the decoder is trained to predict future observations given specified latent actions. We implement the encoder as a spatial-temporal transformer~\cite{xu2020st-trans} with casual temporal masks, following~\citet{villegas2022phenaki}. A group of learnable action tokens $a_{q} \in \mathbb{R}^{N\times d}$, with predefined dimension $d$, are concatenated sequentially to the video features to extract the dynamics. 

To further compress the information and align it with the learning objective~\cite{radford2018improving} of an auto-regressive transformer-based policy, we apply latent quantization to the action tokens. Quantized action tokens $a_{z} \in \mathcal{R}^{N\times d}$ are optimized with VQ-VAE~\cite{van2017neural} objective, with a codebook of $|C|$ vocabulary size. The decoder, implemented as a spatial transformer, is optimized to predict future frames utilizing only the quantized action tokens. We do not feed decoder with historical frames to prevent the model from over-relying on contextual information or merely memorizing the dataset. 

While recent works~\cite{bruce2024genie,gao2025adaworld,ye2024lapa} employs raw pixels for prediction, we observe that pixel-space prediction forces models to attend to noisy, task-irrelevant details (e.g., textures, lighting)~\cite{hafner2019learning}. This issue is amplified in web-scale and crowd-sourced video datasets~\cite{grauman2022ego4d}, where uncontrolled capture conditions introduce further variability. Inspired by joint-embedding predictive architectures (JEPA)~\cite{assran2023ijepa,bardes2023vjepa,zhou2024dinowm}, we propose using DINOv2~\cite{oquab2023dinov2} spatial patch features as semantically rich representations. Their object-centric and spatially aware properties make them ideal not only as inputs but also as prediction targets for latent action models. Our self-supervised objective minimizes the embedding reconstruction error: $\Vert \hat{O}_{t+k} - O_{t+k} \Vert^{2}$. We use $\{O_{t}, O_{t+k}\}$ to represent the DINOv2 feature of paired video frames $\{o_{t}, o_{t+k}\}$. The compact latent action must thus encode the transformation between observations to minimize prediction error.

\paragraph{Latent action decoupling.} As discussed earlier, the actions of the robots are often entangled with irrelevant environmental variations in web-scale videos. To mitigate the unfavorable effect of task-irrelevant dynamics, we incorporate readily available language instructions into the first training stage of latent action model (\Cref{fig:lam} Left). The language inputs are encoded using a pretrained T5 text encoder~\cite{raffel2020exploring} and serve as conditioning signals in the context for both the encoder and decoder. This process can be formally described as:
\begin{equation}
\small
\left\{
\begin{aligned}
& \text{Encode:}\quad \hat{a}_{\textit{TI}} = \mathcal{I} ([O_{t}; O_{t+k} ; a_{\textit{TI}}  ; \ell]),\ \tilde{a}_{\textit{TI}} = \mathbf{VQ}(\hat{a}_{\textit{TI}}), \\
& \text{Decode:}\quad \hat{O}_{t+k} = \mathcal{F} ([O_{t}; \tilde{a}_{\textit{TI}}; \ell]),
\end{aligned}
\right.
\nonumber
\end{equation}
where $[;]$ denotes sequence-wise concatenation, $\mathbf{VQ}$ represents the codebook for vector quantized action representation, and $\ell$ is the instruction embedding from the T5 text encoder. Sending task instructions to the decoder provides high-level semantic guidance regarding the underlying actions. As a result, the quantized latent actions are optimized to encode only the environmental changes and visual details~\cite{zha2024langToken}, omitting higher-level task-relevant information due to the constrained capacity of the codebook~\cite{alemi2016deepib}. This stage establishes a set of latent actions that encapsulate {\color{LightGreen}\textit{task-irrelevant}} information, such as the emergence of new objects, movements of external agents, or camera-induced motion artifacts. These dynamics, while critical for grounding the model in the visual environment, are orthogonal to the specific objectives of the task.

Following this, we repurpose the task-irrelevant codebook and parameters of the latent action model trained in Stage 1 for the following stage (depicted in~\Cref{fig:lam} Right), where the objective is to learn a new set of {\color{LightRed}\textit{task-centric}} latent actions $\hat{a}_{\textit{TC}}$ upon which the policy is trained. In this stage, the model extracts action information through:
\begin{equation}
\small
\left\{
\begin{aligned}
\text{Encode:} \quad &\{\hat{a}_{\textit{TI}},\ \hat{a}_{\textit{TC}}\} = \mathcal{I}([O_{t}; O_{t+k}; a_{\textit{TI}}; a_{\textit{TC}}]), \\
&\tilde{a}_{\textit{TI}} = \mathbf{VQ}(\hat{a}_{\textit{TI}}),\ \tilde{a}_{\textit{TC}} = \mathbf{VQ}_{\textit{TC}}(\hat{a}_{\textit{TC}}),\\
\text{Decode:} \quad &\hat{O}_{t+k} = \mathcal{F}([O_{t}; \tilde{a}_{\textit{TI}}; \tilde{a}_{\textit{TC}}]),
\end{aligned}
\right.
\nonumber
\end{equation}
where $\mathbf{VQ}_{\textit{TC}}$ denotes the newly initialized codebook for learning task-centric dynamics.
Building upon the acquired task-irrelevant representations, we freeze the corresponding codebook, enabling the model to focus on refining and specializing the new set of latent actions. This specialization facilitates the precise modeling of task-related dynamics, such as object manipulation or goal-directed motion trajectories.
The explicit decoupling of latent action representations enhances our generalist policy’s generalization capability across diverse environments and tasks. Compared to naive latent action learning approaches (\textit{e.g.}, LAPA~\cite{ye2024lapa}), training exclusively on task-centric representations yields faster convergence while achieving robust performance, suggesting these latent actions are more informative for subsequent policy learning.

\subsection{Pretraining of Generalist Policy}
\label{sec:step2}
With the latent action model trained in the preceding step, we proceed to label any video frame $o_{t}$ with latent actions $a_{z}$, given $o_{t+k}$. We then employ those labels to develop a generalist policy. To align with~\citet{kim2024openvla}, our generalist policy is built upon the Prismatic-7B~\cite{karamcheti2024prismatic} vision-language model (VLM). The architecture integrates a fused visual encoder derived from SigLip~\cite{zhai2023siglip} and DINOv2~\cite{oquab2023dinov2}, a projection layer to align visual embeddings with the language modality, and the LLaMA-2 large language model (LLM)~\citep{touvron2023llama}. Unlike prior LLM-based generalist policies (\textit{i.e.,} RT-2~\cite{brohan2023rt2} and OpenVLA~\cite{kim2024openvla}) that directly plan in low-level action spaces by mapping infrequently used words in the LLaMA tokenizer vocabulary to uniformly distributed action bins within $[-1,1]$, we extend the vocabulary with $|C|$ special tokens, specifically \{\texttt{ACT\_1}, \texttt{ACT\_2}, \texttt{ACT\_3},..., \texttt{ACT\_C}\}. Latent actions are projected into this vocabulary based on their indices in the action codebook. This approach preserves the original model architecture and training objectives of the VLM, fully leveraging its pretrained knowledge for transfer to robotic control tasks. Specifically, our policy model $\pi_{\phi}$ receives observation $o_{t}$, task instructions $l$ and prefixes of latent action tokens $a_{z, < i}$, and is optimized to minimize the sum of next-latent-action negative log-probabilities:
\begin{equation}
\small
    \mathcal{L} = \mathbb{E}_{o_{t}, l, a_{z, <i}} \left[ - \sum_{i=1}^{N} \log \ \pi_{\phi}(\hat{a}_{z, i} = a_{z, i}\ | \ o_{t}, l, a_{z, < i} ) \right],
    \nonumber
\end{equation}
where $N$ represents the total length of action tokens. We set $N=4$ for all our experiments. Moreover, empirical evidence indicates that a compressed action space (\eg, reducing from $256^{7}$ in OpenVLA~\cite{kim2024openvla} to $16^{4}$ when $|C|=16$) significantly accelerates model convergence. Our approach achieves competitive results with only 960 A100-hours of pretraining, a substantial reduction compared to the 21,500 A100-hours required for OpenVLA pretraining.

By training our policy within a unified latent action space, the model capitalizes on transferable knowledge derived from cross-domain datasets. Unlike~\citet{yang2024pushing} which necessitates manual alignment of action spaces through visually similar egocentric motions, such as wrist camera movements in manipulation tasks and egocentric navigation, our method eliminates this requirement. Consequently, \modelname expands the scope of utilizable datasets and enhances overall performance, demonstrating the efficacy of leveraging task-centric latent action representations for scalable policy learning.

\begin{figure}[t]
    \centering
    \includegraphics[width=0.97\linewidth]{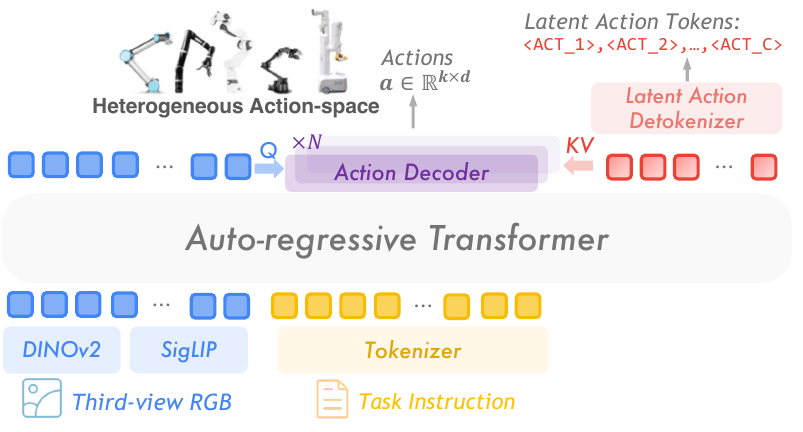}
    \caption{\textbf{Architecture of the generalist policy.} Our policy architecture is founded on the Prismatic-7B Vision-Language Model (VLM)~\cite{karamcheti2024prismatic}, which processes projected visual embeddings and tokenized task instructions as inputs to predict latent action tokens in an auto-regressive manner. To adapt to specific robotic systems, specialized action decoder heads are employed. These decoders leverage visual information to extract context-specific features from latent actions and subsequently translate them into executable control signals of robotic systems with heterogeneous action spaces.}
    \label{fig:policy}
\end{figure}

\begin{figure*}[t]
    \centering
\begin{minipage}{0.405\linewidth}
    \includegraphics[width=\linewidth]{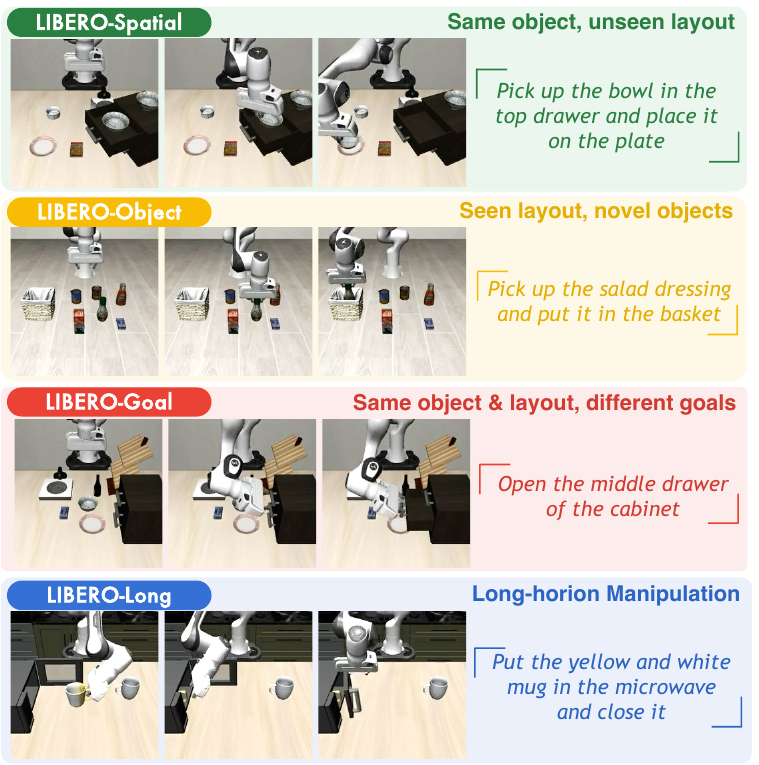}
    \caption{{Task setup on the LIBERO benchmark.}}
    \label{fig:libero_intro}
\end{minipage}
\hspace{4pt}
\begin{minipage}{0.55\linewidth}
    \captionof{table}{\textbf{Results on LIBERO benchmark across four evaluation suites.} Our proposed \modelname exhibits superior performance across all benchmarked tasks compared to existing baseline methods, attributable to its enhanced knowledge transferability and generalization capabilities. Our model achieves state-of-the-art results despite being pretrained exclusively on either the Bridge-V2~\cite{walke2023bridgedata} dataset or action-free human video data (denoted as ``Bridge'' and ``Human'' respectively). $^{\dag}$Methods use additional wrist-view camera inputs. $^{*}$We reproduced results of LAPA using the Prismatic-7B VLM.}
    \small
    \setlength{\tabcolsep}{9pt}
    \scalebox{0.93}{
    \begin{tabular}{l|l|cccc|c}
    \toprule
    \multicolumn{2}{l|}{Method} & Spatial & Object & Goal & Long & Average \\
    \midrule
    \multicolumn{2}{l|}{LAPA$^{*}$~\cite{ye2024lapa}} & 73.8 & 74.6 & 58.8 & 55.4 & 65.7 \\
    \multicolumn{2}{l|}{Diffusion Policy~\cite{chi2023diffusion}} & 78.3 & 92.5 & 68.3 & 50.5 & 72.4 \\
    \multicolumn{2}{l|}{Octo~\cite{team2024octo}} & 78.9 &85.7& 84.6 &51.1 &75.1\\
    \multicolumn{2}{l|}{MDT$^{\dag}$~\cite{reuss2024multimodal}} &78.5 &87.5 &73.5 &64.8 &76.1\\
    \multicolumn{2}{l|}{OpenVLA~\cite{kim2024openvla}} & 84.7 &88.4 &79.2 &53.7 &76.5\\
    \multicolumn{2}{l|}{MaIL$^{\dag}$~\cite{jia2024mail}} & 74.3 &90.1 &81.8 &78.6 &83.5\\
     \midrule
    \multirow{3}{*}{\makecell[c]{\modelname\\(Ours)}} & Human &91.2 &94.2 &90.2 & 79.4 & 88.7\\
     & Bridge &95.2 & 95.4 & 91.9 & 87.5 & 92.5\\ 
     & \baseline{Full} & \baseline{\textbf{96.5}} & \baseline{\textbf{96.8}} & \baseline{\textbf{95.6}} & \baseline{\textbf{92.0}} & \baseline{\textbf{95.2}} \\  
    \bottomrule
    \end{tabular}}
    \label{tab:libero}
\end{minipage}
\end{figure*}

\subsection{Post-training for Deployment}
\label{sec:step3}

\paragraph{Latent action decoding.} During downstream adaptation, the pre-trained generalist policy maintains its embodiment-agnostic characteristics by predicting the next latent action during downstream adaptation. To bridge the gap between latent actions and executable behaviors, additional action decoders are employed (as depicted in~\Cref{fig:policy}). 
Specifically, the sequence of visual embeddings is first aggregated into a single token through multi-head attention pooling~\cite{lee2019set}, which then functions as the query to extract information from the latent action embeddings. This process is formulated as:
\begin{equation}
\small
\left\{
\begin{aligned}
\text{Visual Embed.:}\quad &\textcolor{cvprblue}{E_{v}'} = \mathcal{A}(Q = q_{v}, K=V= \textcolor{cvprblue}{E_{v}}), \\
\text{Action Embed.:}\quad &\textcolor{LightRed}{E_{a}'} = \mathcal{A}(Q = q_{a} + \textcolor{cvprblue}{E_{v}'}, K=V= \textcolor{LightRed}{E_{a}}),
\end{aligned}
\right.
\nonumber
\end{equation}
where $\mathcal{A}$ represents multi-head attention, $\{E_{v}, E_{a}\}$ are visual and latent action embeddings from the last layer of VLM, and $\{q_{v}, q_{a}\}$ are randomly initialized queries to extract visual and action information respectively. The resultant action embedding $E_{a}'$ is subsequently projected linearly into the desired action space of the target robotic system. Given that latent actions are designed to represent actions occurring within approximately a one-second interval (mentioned in~\Cref{sec:step1}), they can be naturally decoded into action chunks~\cite{zhao2023learning}. The chunk size can be easily customized for specific embodiments to achieve smoother and more precise control.

In practice, we employ parameter-efficient fine-tuning using LoRA~\cite{hu2021lora} to achieve efficient adaptation. With the addition of the action head comprising merely 12.6M parameters, the total number of trainable parameters is approximately 123M. The entire model is trained end-to-end, optimizing both the next-latent action prediction loss and the L1 loss between the ground-truth and predicted low-level actions. 

\paragraph{Learn from history outputs.} Historical observations have been demonstrated to play a critical role in enhancing sequential decision-making processes for robotic control~\cite{meuleau2013learning,kurniawati2021partially,li2024robovlm}. However, directly providing large vision-language-action models with multiple historical observations introduces significant inference latency and results in redundant information within visual tokens~\cite{zheng2024tracevla,li2024robovlm}. Drawing inspiration from the well-established Chain-of-Thought (CoT) reasoning paradigm~\cite{wei2022chain} in large language models (LLMs), which generates intermediate reasoning steps to address complex tasks, we propose leveraging historical latent action outputs to facilitate decision-making in robotic control. Much like LLMs resolve questions step-by-step, we incorporate past actions into the input prompt at each timestep during rollouts. This establishes a feedback loop for the robot policy, enabling policy to learn from its own decisions and adapt to dynamic environments. 

To operationalize this approach, we employ the latent action model to annotate actions extracted from historical frames. These annotated actions are then mapped into the LLaMA token vocabulary and appended to task instructions. During post-training, historical action inputs are integrated as inputs to endow the model with in-context learning capabilities. At inference time, one step of historical latent action (encoded as $N=4$ tokens) is incorporated at each timestep, with the exception of the initial step. Empirical results demonstrate that this straightforward design improves model performance, particularly in long-horizon tasks (see \Cref{sec:abl}).

\section{Evaluations}

To demonstrate the performance of our proposed generalist policy, our evaluation framework assesses the capabilities of \modelname across a diverse suite of benchmarks (including manipulation benchmarks: LIBERO~\cite{liu2024libero}, CALVIN~\cite{mees2022calvin}, SimplerEnv~\cite{li2024simplerenv}, and a navigation benchmark: R2R~\cite{anderson2018r2r}) and real-world scenarios. Additionally, we conduct latent action analysis to quantify the task-centric property, and perform ablation studies to explore critical design choices. With comprehensive evaluations, we mainly intend to investigate:

\begin{enumerate}
    \item \textbf{Performance \& Adaptability.} Can \modelname successfully transfer the knowledge acquired during pretraining to novel embodiments and tasks and adapt efficiently? (See \Cref{sec:manip} for manipulation performance and \Cref{sec:navi} for adaptability to navigation.)
    \item \textbf{Generalizability.} How does \modelname generalize to unseen scenarios? (See \Cref{sec:real-world} for the analysis of its generalizability in novel settings.)
    \item \textbf{Scalability.} Can \modelname effectively utilize diverse data sources, even including human videos, and derive scalable benefits from the continuously expanding dataset? (See~\Cref{sec:abl} for data scalability analysis.)
\end{enumerate}

\begin{figure*}[t]
    \centering
    \includegraphics[width=0.97\linewidth]{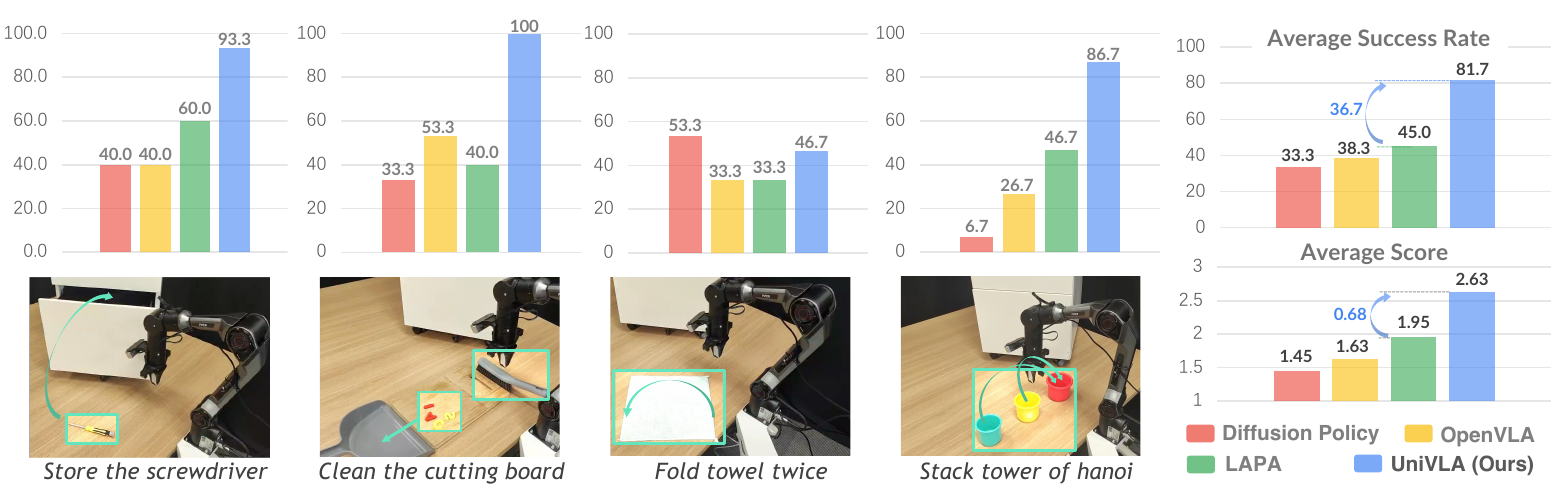}
    \caption{\textbf{Real-world robot experiments.} We propose four different tasks: ``Store the screwdriver'', ``Clean the cutting board'', ``Fold towel twice'', and ``Stack tower of hanoi'', towards the evaluation of four axis of policy's capabilities. \modelname outperforms previous state-of-the-art with an average elevation of 36.7\% success rate and 0.68 average score across all tasks.}
    \label{fig:real-world}
    \vspace{-6pt}
\end{figure*}

\begin{figure}[t]
    \centering
    \includegraphics[width=0.95\linewidth]{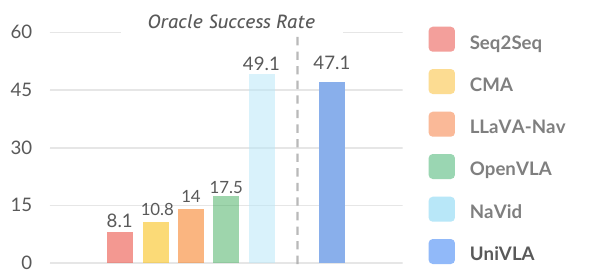}
    \caption{\textbf{Oracle success rate on R2R in VLN-CE.} With only a single-frame RGB input, \modelname demonstrates performance on par with NaVid, a navigation model that incorporates the entirety of historical observations, while markedly outperforming OpenVLA in success rate.
    }
    \vspace{-6pt}
    \label{fig:navigation-result}
\end{figure}

\subsection{Main Results}
\label{sec:manip}

\subsubsection{Manipulation Benchmark on LIBERO} 

\paragraph{Experiment setup.} We pretrain our full latent action model on manipulation data, navigation data and human videos data, which are a subset of Open X-Embodiment (OpenX) dataset~\cite{padalkar2023open}, GNM dataset~\cite{shah2023gnm}, and human videos (Ego4D \cite{grauman2022ego4d}) respectively. The pretraining details can be found in \Cref{sec:Extra_Pretraining_Details}. The LIBERO benchmark~\cite{liu2024libero} comprises four task suites specifically designed to facilitate research on lifelong learning in robotic manipulation. Our experiments exclusively focus on supervised fine-tuning within the target task suite, evaluating the performance of various policies trained through behavioral cloning on successful task demonstrations. As illustrated in~\Cref{fig:libero_intro}, our experimental setup includes the following task suites, each consisting of 10 tasks with 50 human-teleoperated demonstrations per task: 
\begin{enumerate}
\item \textbf{LIBERO-Spatial} requires the policy to infer spatial relationships to accurately place a bowl, evaluating the model's ability to reason about geometric configurations;
\item \textbf{LIBERO-Object} maintains identical scene layouts but introduces variations in object types, assessing the policy's capacity to generalize across object instances;
\item \textbf{LIBERO-Goal} retains consistent objects and layouts while assigning diverse task objectives, challenging the policy to exhibit goal-oriented behavior and adaptability;
\item \textbf{LIBERO-Long} focuses on long-horizon manipulation tasks involving multiple sub-goals, incorporating heterogeneous objects, layouts, and task sequences to evaluate the model's proficiency in complex, multi-step planning.
\end{enumerate}

We adhere to the data processing pipeline introduced in OpenVLA~\cite{kim2024openvla} to exclude failure cases from the demonstration data used for training. \modelname is trained on LIBERO-Long for 40k steps and other test suites for 30k steps, with a global batch size of 128. We only use third-person image and language instructions as inputs. Notably, \textit{none} of the samples in LIBERO is included in the pretraining dataset of policy, and the training data for our latent action model, necessitating generalizability for both. In addition to presenting the results of our most performant model, which is pre-trained on the full dataset, we also provide results from models pre-trained exclusively on the Bridge-V2~\cite{walke2023bridgedata} and human data, denoted as ``Bridge'' and ``Human'' in~\Cref{tab:libero}, respectively. To minimize variance, all methods are evaluated over 500 trials per task suite (\textit{i.e.,} 50 trials per task), with the reported performance reflecting the average success rate across three seeds.

\paragraph{Baselines.}
Our selected baseline models include the following five representative models, where OpenVLA and LAPA are more closely related to our method:
\begin{itemize}
\item \textbf{LAPA}~\cite{ye2024lapa} introduces an unsupervised framework for learning latent actions from unlabeled human videos. 

\item \textbf{Octo}~\cite{team2024octo} is a transformer-based policy trained on diverse robotic datasets, which employs a unified action representation to handle heterogeneous action spaces.

\item \textbf{MDT}~\cite{reuss2024multimodal} leverages diffusion models to generate flexible action sequences conditioned by multimodal goals. 

\item \textbf{OpenVLA}~\cite{kim2024openvla} is a vision-language-action model that leverages large-scale pretraining on diverse datasets, including OpenX, to enable generalist robotic policies. 

\item \textbf{MaIL}~\cite{jia2024mail} enhances imitation learning by incorporating selective state space models, which improve the efficiency and scalability of policy learning. 
\end{itemize}

\paragraph{Results.} The results presented in~\Cref{tab:libero} demonstrate the exceptional performance of UniVLA across all four evaluation suites, significantly outperforming prior generalist policies such as OpenVLA, LAPA, and Octo. Notably, UniVLA achieves an average performance of 95.2\% by pretraining on the full dataset, surpassing OpenVLA and LAPA by margins of 18.7\% and 29.5\% respectively. Despite being pretrained solely on the Bridge-V2 dataset, \modelname attains 92.5\% average performance, outperforming methods like MaIL (83.5\%) and MDT (76.1\%) that leverage additional wrist-view camera inputs. Pretraining our policy with human data outcompetes OpenVLA, which is trained with in-domain OpenX data, by a margin of 12.2\%. In conclusion, UniVLA shows unparalleled knowledge transfer capability and establishes a new state-of-the-art on LIBERO benchmark. We provide additional results on CALVIN and SimplerEnv benchmark in~\Cref{sec:additional_results}.

\begin{table*}[t!]
    \centering
    \caption{\textbf{Generalizability evaluation.} \modelname demonstrates superior performance across all evaluated tasks, showcasing its exceptional ability to generalize from high-level semantic comprehension to low-level visual robustness.}
    \label{tab:generalizability}
    \small
    \setlength{\tabcolsep}{5.5mm}
    \scalebox{0.91}{
    \begin{tabular}{l|cccccc|cc}
    \toprule
          \multirow{2}{*}{Method} &  \multicolumn{2}{c}{\makecell[c]{Lightning Variation}} & \multicolumn{2}{c}{\makecell[c]{Visual Distractor}} & \multicolumn{2}{c|}{\makecell[c]{Novel Object}} & \multicolumn{2}{c}{Average $\uparrow$}\\
         &Succ. & Score &Succ. & Score &Succ. & Score &Succ.  & Score \\
    \midrule
    Diffusion Policy~\cite{chi2023diffusion} & 20.0 & 0.60 & 26.7 & 0.80 & 26.7 & 0.67 & 24.4 & 0.69\\
    OpenVLA~\cite{kim2024openvla} & 13.3 & 0.93 & 20.0 & 0.73 &26.7 & 1.27 & 20.0 & 0.98\\
    LAPA~\cite{ye2024lapa} & 26.7 & 1.60 & 6.7 & 0.6 & 53.3 & 1.87 & 28.9 & 1.36\\
    \midrule
     \baseline{\modelname (Ours)}    & \baseline{\textbf{66.7}} & \baseline{\textbf{2.33}} & \baseline{\textbf{53.3}} & \baseline{\textbf{2.40}} & 
     \baseline{\textbf{86.7}} & \baseline{\textbf{2.73}} & \baseline{\textbf{68.9}} & \baseline{\textbf{2.49}}\\
 
    \bottomrule
    \end{tabular}}
\end{table*}

\medskip
\subsubsection{Navigation Benchmark on Room2Room}
\label{sec:navi}
\paragraph{Experiment setup.}
In this experiment, we evaluate \modelname on the VLN-CE benchmarks~\cite{krantz2020vlnce} to assess its performance on navigation tasks. These benchmarks offer a set of language-guided navigation tasks and continuous environments for executing low-level actions in reconstructed photorealistic indoor scenes. Specifically, we focus on the Room2Room (R2R)~\cite{anderson2018r2r} task in VLN-CE, one of the most widely recognized benchmarks in vision-and-language navigation (VLN). All methods are trained on the 10,819 samples in the R2R training split and evaluated on the 1,839 samples in the R2R val-unseen split. 
We use the oracle success rate to evaluate navigation performance. An episode is considered successful if the agent arrives within 3 meters of the goal in the VLN-CE.

\paragraph{Baselines.}
To ensure a fair comparison with \modelname, we evaluate RGB-only methods that operate without depth or odometry data, directly predicting low-level actions within the VLN-CE environments. Selected baselines are as follows:
\begin{itemize}
    \item \textbf{Seq2Seq}~\cite{Krantz2020BeyondTN} is a recurrent sequence-to-sequence policy that predicts actions from RGB observations.
    \item \textbf{CMA}~\cite{Krantz2020BeyondTN} employs cross-modal attention to integrate instructions with RGB observations for action prediction.
    \item \textbf{LLaVA-Nav} is a modified version of LLaVA~\cite{Liu2023LLaVA}, co-finetuned with data proposed by NaVid~\cite{zhang2024navid}, and encodes history using an observation-to-history technique.
    \item \textbf{OpenVLA}~\cite{kim2024openvla} is a vision-language-action model. We introduce several special tokens to tokenize navigation actions and finetune the model on the R2R training split.
    \item \textbf{NaVid}~\cite{zhang2024navid} is a video-based large vision-language model that encodes all historical RGB observations. It uses a pretrained vision encoder to encode visual observations and a pretrained LLM to predict actions.
\end{itemize}

\paragraph{Results.} In~\Cref{fig:navigation-result}, we report the oracle success rate for each method. \modelname significantly outperforms Seq2Seq and CMA, increasing the oracle success rate from 8.10\% to 47.1\%. Given the high computational cost of prompting history in LLaVA-Nav, we refer to NaVid and present its results on a 100-episode subset of the VLN-CE R2R val-unseen split. \modelname surpasses the oracle success rate of LLaVA-Nav by \textbf{33.1\%} and OpenVLA by \textbf{29.6\%}. Furthermore, \modelname achieves an oracle success rate comparable to NaVid, which encodes all historical observations, while \modelname conditions only on the current observation and historical latent action.

\begin{figure}[t]
    \centering
    \includegraphics[width=0.97\linewidth]{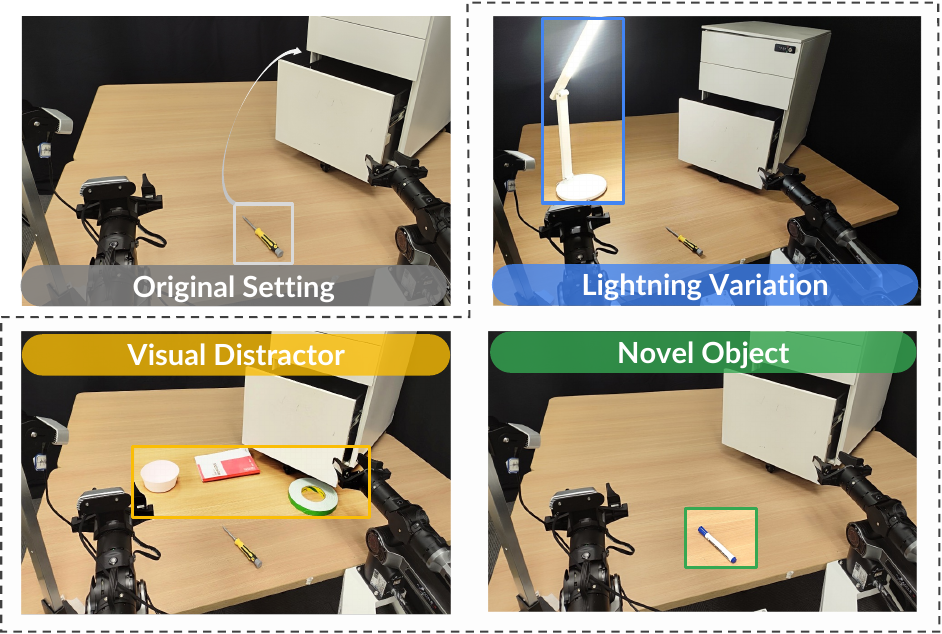}
    \caption{\textbf{Setting on generalizability evaluations.} We evaluate the generalizability of policies in 3 different settings. \textbf{(a)} Lightning Variation: We dimmed the ambient light and applied strong lighting in a specified direction. \textbf{(b)} Visual Distractor: We added a bowl, notebook, and tape on the tabletop. \textbf{(c)} Novel Object: We replaced the object to be manipulated from a screwdriver to an unseen marker pen.}
    \vspace{-5pt}
    \label{fig:real-world-gen}
\end{figure}

\medskip
\subsubsection{Real-world Robot Deployment}
\label{sec:real-world}

\paragraph{Experiment setup.} All real-world experiments are conducted with a Piper arm from AgileX Robotics featuring a 7-DoF action space and a third-view Orbecc DABAI RGB-D camera, which we only utilize RGB images as input. To evaluate policies, we design a comprehensive set of tasks that span various dimensions of policy capabilities, including: 
\begin{enumerate} 
    \item \textbf{Spatial Awareness:} Pick up the screwdriver to put it into the cabinet and close the door (``Store the screwdriver''). 
    \item \textbf{Tool-usage and Nonprehensile Manipulation:} Pick up the broom and sweep the items on the cutting board into the dustpan (``Clean the cutting board'').
    \item \textbf{Deformable Objects Manipulation:} Fold the towel in half twice (``Fold towel twice'').
    \item \textbf{Semantic Understanding:} Stack the medium tower on top of the large one first, then stack the small one on top of the medium one.(``Stack tower of hanoi'')
\end{enumerate}

For each task, we collect 20–80 trajectories, scaled according to task complexity, to finetune our model. To evaluate generalization comprehensively, we design experiments that span multiple axes of unseen scenarios, including lighting variations, visual distractors, and object generalization (see~\Cref{fig:real-world-gen}). Recognizing that success rate alone inadequately captures policy performance or distinguishes their capabilities, we introduce a step-wise scoring system. For each of the four tasks, we assign a maximum score of 3 points, reflecting the completion of distinct stages during task execution. Detailed scoring criteria, task setup and experiment results are provided in~\Cref{sec:details_real_robots}.

\begin{figure*}[t]
    \centering
    \includegraphics[width=0.97\linewidth]{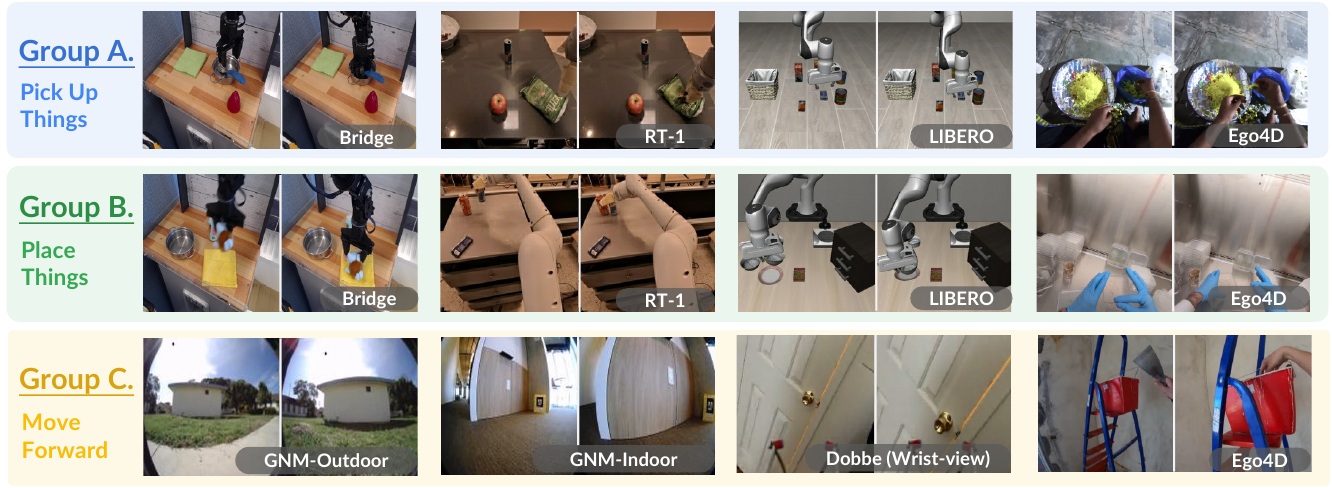}
    \caption{\textbf{Latent action analysis.} We plot image pairs labeled with the same latent action from different sources of data and embodiments. Each group of latent actions exhibits semantic-consistent actions. 
    More examples are in \Cref{sec:additional_results}.}
    \label{fig:latent_action}
\end{figure*}

\begin{figure}[t]
    \centering
    \includegraphics[width=0.92\linewidth]{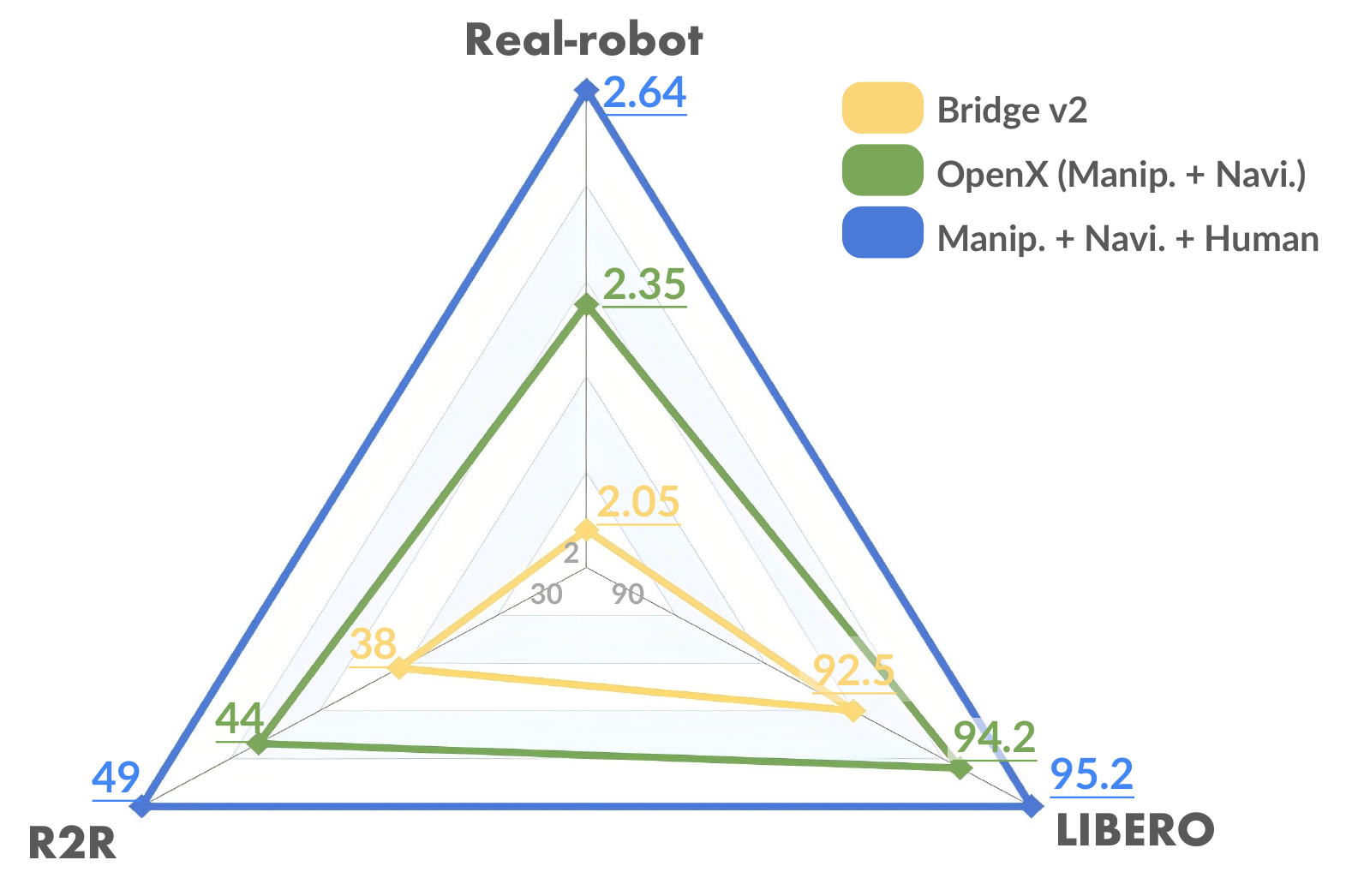}
    \caption{\textbf{Data scalability.}  \modelname effectively expands its pretraining corpus by incorporating cross-embodiment data from OpenX and unlabeled human demonstrations, leading to continuously improved downstream performance.}
    \vspace{-5pt}
    \label{fig:dara_scale}
\end{figure}

\paragraph{Baselines.} We choose Diffusion Policy~\cite{chi2023diffusion}, alongside generalist policies, OpenVLA~\cite{kim2024openvla} and LAPA~\cite{ye2024lapa} as our baselines. Diffusion Policy is trained in a single-task manner, whereas the generalist models are trained on all tasks simultaneously with instruction inputs. For a fair comparison, we reproduce LAPA with Prismatic-7B VLM~\cite{karamcheti2024prismatic} and action decoder heads, aligning its architecture with our method. This setup allows us to isolate and emphasize the contribution of our task-centric latent action space. Specific parameters and architectural details can be found in \Cref{sec:details_real_robots}.

\paragraph{Results.} We plot task  success rates in \Cref{fig:real-world}. The single-task Diffusion Policy (DP), optimized for trajectory fidelity and low-latency control, excels in tasks like towel folding, where success hinges on executing a fixed trajectory once the correct towel edge is selected. This specialization allows DP to achieve a higher success rate (53.3\%) compared to \modelname (46.7\%) in this task. However, \modelname achieves a higher step-wise score (2.47 vs. DP’s 2.33, detailed in~\Cref{sec:details_real_robots}), reflecting its ability to reliably complete intermediate stages (e.g., edge selection, partial folding) even when final execution falters—a critical advantage in dynamic real-world environments where partial progress is valuable.

This trade-off arises from UniVLA’s generalist design: while DP’s single-task training maximizes trajectory precision for specific workflows, it struggles in tasks requiring semantic reasoning (\textit{e.g.}, stack tower of hanoi, where DP achieves only 6.7\% success). In contrast, \modelname demonstrates superior generalization and semantic understanding, achieving an unparalleled 86.7\% success rate. This is further evidenced by a 93.3\% success rate in scenarios requiring precise object manipulation and spatial reasoning (where the object is placed at varied poses and positions in ``Store the screwdriver'' task).

In addition, our method achieves a real-time, closed-loop inference frequency of 10Hz on an NVIDIA RTX 4090 GPU by planning in a compact latent action space, and allowing efficient action chunk prediction (we use a chunk size of 12 in practice). OpenVLA, despite extensive training on large-scale robot datasets, suffers from execution stuttering due to inference latency (\textit{e.g.}, 0.18s when predicting a single action step, 0.68s when predicting action chunks with size 4), resulting in poor real-world performance with only a 38.3\% average success rate. In a nutshell, \modelname outperforms LAPA, the second best policy, by \textbf{36.7\%} in success rate and \textbf{0.68} in average score, demonstrating its real-world effectiveness and the advantages of our proposed task-centric latent action space.

\paragraph{Generalizability Analysis.} We investigate the generalizability of policies from 3 different aspects, with the specific experiment setups shown in \Cref{fig:real-world-gen}. 
The results in~\Cref{tab:generalizability} highlight UniVLA’s exceptional generalizability, significantly outperforming baseline methods in success rates and step-wise scores. It achieves a 66.7\% success rate under varying lighting conditions, surpassing Diffusion Policy (20.0\%), OpenVLA (13.3\%), and LAPA (26.7\%), demonstrating robustness to environmental change. In scenarios with visual distractors, policies that rely more on semantic information, such as LAPA and UniVLA, experience a relatively notable performance drop. In the novel object setting, we replaced the screwdriver with a marker and adjusted the language inputs for the generalist policy accordingly. This change had minimal impact on our policy as the success rate only drops by 6.6\%. Overall, UniVLA achieves an average success rate of \textbf{68.9\%} and an average score of \textbf{2.49}, significantly outperforming prior VLAs like LAPA (28.9\%, 1.36) and OpenVLA (20.0\%, 0.98). We also provide video demos in the supplementary material.

\subsection{Discussion on Latent Action}
\label{sec:LAM_analysis}

\paragraph{Qualitative analysis.} We investigate the cross-domain transferability of latent actions by visualizing image pairs from different data sources sharing the same latent action in \Cref{fig:latent_action}. Each group of latent actions maps to semantically consistent behaviors across embodiments (\textit{e.g.,} latent actions representing ``\texttt{Pick up things}'' in {\color{DeepBlue}Group A}). Notably, our latent action model, trained without any data from the LIBERO~\cite{liu2024libero} dataset, generalizes effectively to label accurate actions in this unseen domain. Furthermore, LAM learns to align wrist-view observations in manipulation with ego-centric movements in navigation, as demonstrated in {\color{yellow}Group C}, highlighting its ability to bridge diverse modalities and embodiments.

\paragraph{Quantitative analysis.} To evaluate the effectiveness of our proposed dynamics decomposition approach for task-centric latent action learning, we assess the deployment performance of policies trained with labels derived from different latent actions. The results on LIBERO are shown in~\Cref{table:human_only_comparison}. We pre-train policies using only human videos, which contain significant amounts of unpredictable motion, to amplify the advantages of our method. In comparison to the latent action construction approach introduced in Genie~\cite{bruce2024genie}, which captures all visual changes, our method demonstrates clear superiority. Specifically, we achieve a 6.4\% improvement in average success rate, with substantial gains in LIBERO-Goal and LIBERO-Long (13\% and 9.8\% improvement, respectively). In contrast, latent actions that are task-irrelevant are poorly aligned with true actions, making it difficult for policies to infer them from observations and task instructions. This is reflected in both lower action token prediction accuracy during training and poorer inference performance. Notably, training with task-irrelevant latent actions results in near-zero success rates on the challenging LIBERO-Long benchmark.

\begin{table}[tb!]
  \small
  \centering
  \caption{
  \textbf{Performance on LIBERO using various latent actions.} We pretrain policies using different latent actions on Ego4D~\cite{grauman2022ego4d}, which features human videos with diverse movements and task-irrelevant dynamics, to demonstrate our successful decoupling of task-centric dynamics. While task-irrelevant ones yield poor performance, task-centric latent action learning produces more meaningful action representations, ultimately achieving superior deployment success rates.
  }
  \label{table:human_only_comparison}
  \setlength{\tabcolsep}{3.5mm}
  \scalebox{0.9}{
  \begin{tabular}{l| c c c c | c}
    \toprule
         Latent Action
         & Spatial & Object & Goal & Long & Avg.\\
         \midrule
        Genie~\cite{bruce2024genie} &89.8 &92.8 & 77.2 & 69.6 & 82.3\\
        \midrule
        Task-irrelevant &68.0 &90.4 & 67.2 & 0.2 & 56.5 \\
        \baseline{Task-centric} &\baseline{\textbf{91.2}} &\baseline{\textbf{94.2}}  & \baseline{\textbf{90.2}} & \baseline{\textbf{79.4}} & \baseline{\textbf{88.7}} \\
    \bottomrule
    \end{tabular}
}
\end{table}

\begin{figure}[t]
    \centering
    \includegraphics[width=\linewidth]{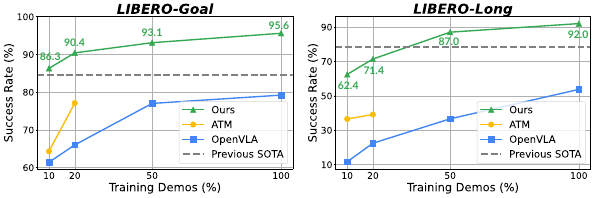}
    \caption{\textbf{Data efficiency.} We present the success rate of \modelname across varying dataset proportions (10\%, 20\%, 50\%, and the full dataset). Our policy can be adapted to an unseen environment without requiring extensive expert demos for training, showing notable superiority over baselines.}
    \label{fig:data_effi}
\end{figure}

\begin{table}[t]
  \small
  \centering
  \caption{
  \textbf{Ablations on decoder design.} ``Auto-regressive'' represents that we follow the approach of OpenVLA and LAPA, predicting actions sequentially over discretized action bins in an auto-regressive fashion. ``w/o visual'' indicates that visual embeddings are not utilized as query inputs for decoding latent actions, as depicted in~\Cref{fig:policy}. The proposed action decoder head, augmented by visual features, proves to be the most effective, yielding the highest results on all test suites.
  }
  \label{table:decoder_design}
  \setlength{\tabcolsep}{3mm}
  \scalebox{0.9}{
  \begin{tabular}{l| c c c c | c}
    \toprule
         Action Decoder & Spatial & Object & Goal & Long & Avg. \\
         \midrule
        Auto-regressive & 85.2 & 81.2 & 79.0 & 49.0 & 73.6\\
        \midrule
        Ours w/o Visual & 95.0 & 95.4 & 93.7 & 86.0 & 92.5\\
        \baseline{Ours} & \baseline{\textbf{96.5}} &\baseline{\textbf{96.8}} & \baseline{\textbf{95.6}} & \baseline{\textbf{92.0}} & \baseline{\textbf{95.2}} \\
    \bottomrule
    \end{tabular}
    }
\end{table}

\begin{table}[t]
  \small
  \centering
  \caption{
  \textbf{Ablations on the use of history action.} Incorporating latent action outputs from previous steps as prompt inputs, despite its simplicity, enhances performance, particularly in long-horizon tasks, such as LIBERO-Long and R2R.
  }
  \label{table:history_latent_action}
  \setlength{\tabcolsep}{4.4mm}
  \scalebox{0.9}{
  \begin{tabular}{l| c c | c }
    \toprule
         \multirow{2}{*}{Prompt Input} &  \multicolumn{2}{c|}{LIBERO (Manip.)}  & \multirow{2}{*}{R2R (Navi.)}\\
         & Goal & Long & \\
         \midrule
        Instruction-only &95.0 &88.1 & 30.6\\
        \baseline{w/ History Action} & \baseline{\textbf{95.6}} & \baseline{\textbf{92.0}} & \baseline{\textbf{47.1}}\\
    \bottomrule
    \end{tabular}
}
\end{table}

\subsection{More Ablations}
\label{sec:abl}

\paragraph{Data scalability.} We show how \modelname evolves with the growing data scale and the incorporation of data from distinct domains in~\Cref{fig:dara_scale}. Though \modelname already sets a new state-of-the-art on LIBERO by pretraining only with Bridge-V2~\cite{walke2023bridgedata}. Cross-embodiment data in OpenX~\cite{padalkar2023open} and Ego4D~\cite{grauman2022ego4d} further amplifies the average success rate by 2.0\%. While the performance on the LIBERO benchmark appears to plateau, consistent performance improvements are observed across our challenging real-world test suites. In real-world evaluations, expanding the pretraining data to OpenX increases the average score by 0.3, compared to Bridge-only pretraining. Further incorporating human data, despite the absence of action labels and the substantial embodiment gap it introduces, yields an additional 0.28 increase. This trend of performance improvement is similarly observed in the R2R navigation benchmark, highlighting the scalability of our approach as it effectively leverages diverse data sources.

\paragraph{Data efficiency.} The preceding section highlights \modelname's scalability with respect to pretraining data, consistently enhancing its capabilities. We next explore its ability to adapt efficiently to unseen environments with minimal data, as detailed in~\Cref{fig:data_effi}. Specifically, we evaluate performance on the LIBERO-Goal and LIBERO-Long benchmarks using partial training data. \modelname demonstrates superior data efficiency compared to prior generalist policy, such as OpenVLA~\cite{kim2024openvla}, and explicit point prediction methods like ATM~\cite{wen2023any}. Notably, with only 10\% of the demonstration data, \modelname achieves a higher success rate on LIBERO-Goal (86.3\% \textit{vs.} 79.2\%) than OpenVLA trained on the full dataset. Moreover, it sets a new state-of-the-art performance on both LIBERO-Goal and LIBERO-Long with only 10\% and 50\% of the training episodes, respectively. By planning within a unified latent action space, \modelname maximizes pretraining knowledge, enabling highly efficient adaptation to new environments.

\paragraph{Latent action decoder.} We compare our proposed action decoding scheme with the auto-regressive approach, which sequentially generates discretized actions as in OpenVLA and LAPA. As shown in~\Cref{table:decoder_design}, our method consistently achieves higher success rates across all test suites, with a striking 42.1\% improvement in LIBERO-Long. Leveraging visual embeddings as queries enhances action decoding by reducing ambiguity in the multimodal distribution, yielding an additional 2.2\% gain in average success rate.

Furthermore, as discussed in~\Cref{sec:step1}, latent actions are designed to encapsulate dynamics over a one-second time horizon. Given this temporal structure, decoding latent actions as action chunks~\cite{zhao2023learning} is an intuitive choice, aligning the chunk size with the control frequency of the target embodiment. This is achieved by simply expanding the output dimension of the final linear projection layer, while introducing negligible additional inference cost compared to the auto-regressive approach.

\paragraph{History latent actions.} As detailed in~\Cref{sec:step3}, we augment the instruction input with historical latent actions to enhance sequential decision-making. We evaluate the efficacy of this minimal architectural modification on manipulation and navigation tasks, with quantitative results in~\Cref{table:history_latent_action}. The approach proves particularly impactful in long-horizon scenarios: using only four input tokens (representing one latent action group) improves success rates by 16.5\% (R2R) and 3.9\% (LIBERO-Long). Extending the history horizon yields diminishing returns. Unlike methods requiring redundant multi-frame visual tokens for temporal context (\eg,~\cite{zhang2024navid,li2024robovlm}), our design provides compact historical guidance while enabling iterative policy refinement through self-referential outputs. This streamlined integration enhances contextual awareness without incurring unnecessary computational overhead.

\section{Conclusion} 
\label{sec:conclusion}

In this work, we introduce \modelname, a vision-language-action model that plans within a unified, task-centric latent action space, enabling efficient adaptation to novel robotic setups. Through extensive evaluations, we demonstrate that \modelname establishes state-of-the-art performance across multiple manipulation and navigation benchmarks. The model also exhibits scalability with heterogeneous pretraining data to enhance its downstream performance, and remains highly adaptable even in data-limited scenarios.
We aim for our work to pave the way for the next generation of generalist policies, capable of leveraging web-scale video data for training, regardless of embodiment gaps or the availability of action labels.

\section{Limitations and Future Work} 

\paragraph{Latent action design.} While \modelname advances generalist robotic policies, several limitations remain. The fixed granularity of the latent action and the predefined codebook size may not be optimal for all tasks or embodiments. Exploring adaptive mechanisms to dynamically adjust these based on environmental conditions could potentially improve performance. In addition, \modelname is primarily evaluated on single-arm manipulation tasks. The action granularity represented by latent action tokens are relatively fixed within our framework. Extending the framework to dual-arm humanoid systems or dexterous hands could require more complex and finer-grained action space modeling. We leave this for future exploration.

\paragraph{Requirements on language annotation.} Regarding language granularity, task-relevant latent actions are designed to encode ego-agent movements critical for task completion, while excluding non-ego dynamics (\textit{e.g.}, steam rising from a kettle during ``boiling water''). The majority of our dataset comprises fine-grained instructions that describe short-horizon actions rather than high-level goals. While more expressive language instructions could potentially reduce ambiguity in latent action learning, we want to emphasize that our approach enables scalable learning from instructions of varying granularity. Without any special handling of instruction, our method outperforms naive latent action learning approaches.

\paragraph{Integration with world model.} The decoder of latent action model is essentially a world model, predicting future observations given latent actions. It can be conditioned on latent actions sampled by our policy on-the-fly and generate multiple corresponding visual plans. This opens the door to reference alignment~\cite{zhang2024grape} with reinforcement learning and test-time scaling through planning trees~\cite{du2024video}, where VLMs~\cite{ma2024vision} or heuristic functions can be adopted as reward models.

\paragraph{In-context learning} capability is critical for enhancing the performance ceiling of vision-language-action models. Given our finding that the proposed latent action model can extract transferable motion representations bridging human and robotic manipulations, we propose encoding human demonstration videos into a sequence of compact latent action embeddings, serving as in-context samples (conceptually, the latent action model functions as a video tokenizer). This approach enables zero-shot skill acquisition without additional fine-tuning. We will explore this direction in future work.

\section*{Acknowledgment}
We thank Li Chen, Modi Shi, and Chengen Xie for their valuable feedback and fruitful discussions.
{
\bibliographystyle{plainnat}
\bibliography{bibliography_short, bibliography_custom}
}

\clearpage
\newpage

\title{
\huge{Learning to Act Anywhere with Task-centric Latent Actions} \\
\textit{Supplementary Material}
}

\renewcommand{\thetable}{A-\Roman{table}}
\renewcommand{\thefigure}{A-\arabic{figure}}
\setcounter{figure}{0}
\setcounter{table}{0}  
\setcounter{section}{0}

\appendix

\subsection{Implementation Details}
\label{sec:Implementation_Details}

\subsubsection{Pretraining Details}
\label{sec:Extra_Pretraining_Details}

For robotic manipulation data, we select a subset in Open X-Embodiment  dataset~\cite{padalkar2023open} with single arm end-effector control. For navigation data, we use a sub-split of GNM~\cite{shah2023gnm} dataset containing both indoor and off-road scenes featuring a ego-view fisheye camera. While actions and proprioceptive states are available in the robot datasets, these are excluded during pretraining; only episode frames and text instructions are used. Additionally, we incorporate open-world human videos, specifically ego-centric videos that depict daily human activities from the Ego4D dataset~\cite{grauman2022ego4d}. Notably, with the exception of the SimplerEnv benchmark~\cite{li2024simplerenv}, which is designed to replicate the environmental setup of the Bridge-V2 dataset, \textit{none} of the downstream evaluation environments have been seen by either our policy or the latent action model during pretraining. This necessitates strong generalization capabilities for both. The detailed composition of the datasets and mixture weights are listed in~\Cref{tab:data_mix}.

\begin{table}[th]
    \centering
    \small
       \begin{tabular}{lr}
        \toprule
        \multicolumn{2}{c}{\textbf{Training Dataset Mixture}}\\
        \midrule
        Fractal~\cite{brohan2022rt1} & 13.9\% \\
        Kuka~\cite{kalashnikov2018scalable} & 6.3\% \\
        Bridge~\cite{walke2023bridgedata} & 6.8\% \\
        Taco Play~\cite{mees2023grounding} & 3.5\% \\
        Jaco Play~\cite{dass2023jacoplay} & 0.6\% \\
        Berkeley Cable Routing~\cite{luo2023multistage} & 0.3\% \\
        Roboturk~\cite{ajay2018roboturk} & 2.8\% \\
        Viola~\citep{zhu2023viola} & 1.1\% \\
        Berkeley Autolab UR5~\citep{BerkeleyUR5Website} & 1.4\% \\
        Toto~\citep{zhou2023train} & 2.4\% \\
        Language Table~\cite{lynch2023interactive} & 5.2\% \\
        Stanford Hydra Dataset~\cite{belkhale2023hydra}  & 5.3\% \\
        Austin Buds Dataset~\cite{zhu2022bottom}  & 0.3\% \\
        NYU Franka Play Dataset~\cite{cui2022play}  & 1.0\% \\
        Furniture Bench Dataset~\cite{heo2023furniturebench}  & 2.9\% \\
        UCSD Kitchen Dataset~\cite{ucsd_kitchens}  & \texttt{<}0.1\% \\
        Austin Sailor Dataset~\citep{nasiriany2022sailor}  & 2.6\% \\
        Austin Sirius Dataset~\cite{liu2022robot}  & 2.0\% \\
        DLR EDAN Shared Control~\cite{quere_shared_2020}  & 0.1\% \\
        IAMLab CMU Pickup Insert~\cite{saxena2023multiresolution}  & 1.1\% \\
        UTAustin Mutex~\cite{shah2023mutex} & 2.6\% \\
        Berkeley Fanuc Manipulation~\cite{fanuc_manipulation2023} & 0.9\% \\
        CMU Stretch~\cite{mendonca2023structured} & 0.2\% \\
        BC-Z~\cite{jang2022bc} & 8.8\% \\
        FMB Dataset~\cite{luo2024fmb}  & 8.4\% \\
        DobbE~\cite{shafiullah2023dobbe}  & 1.7\% \\
        \midrule
        RECON~\cite{khazatsky2024droid}  & 8.9\%\\
        CoryHall~\cite{khazatsky2024droid}  & 2.3\%\\
        SACSoN~\cite{khazatsky2024droid}  & 3.5\%\\
        \midrule
        Ego4D~\cite{khazatsky2024droid}  & 3.0\%\\
                \bottomrule
        \end{tabular}%
        \caption{\modelname training data mixture using datasets from the OXE~\cite{padalkar2023open}, GNM~\cite{shah2023gnm} and Ego4D~\cite{grauman2022ego4d}.}
        \label{tab:data_mix}
\end{table}

\begin{table}[t]
    \centering
    \caption{\textbf{Language-conditioned visuomotor control on CALVIN ABC$\rightarrow$D.} We report success rates along with the average length of completed tasks (out of the whole 5 tasks) per evaluation sequence. UniVLA achieves competitive results while being the only method that relies on solely third-view RGB inputs. $^{*}$Reproduced with action chunks prediction. }
    \vspace{-4pt}
    \label{tab:calvin}
    \small
    \setlength{\tabcolsep}{8pt}
    \scalebox{0.89}{
    \begin{tabular}{l|l|ccccc|c}
    \toprule
    \multicolumn{2}{l|}{\multirow{2}{*}{Method}} & \multicolumn{5}{c|}{ Task completed in a row (\%) $\uparrow$} & \multirow{2}{*}{\makecell[c]{Avg.\\Len.}} \\
    \multicolumn{2}{l|}{} & 1 & 2 & 3 & 4 & 5 & \\
    \midrule
    \multicolumn{2}{l|}{RT-1~\cite{brohan2022rt1}}  & 53.3 & 22.2 & 9.4 & 3.8 & 1.3 & 0.90 \\
    \multicolumn{2}{l|}{RoboFlamingo~\cite{li2023roboflamingo}}  & 82.4 & 61.9 & 46.6 & 33.1 & 23.5 & 2.48 \\
    \multicolumn{2}{l|}{SuSIE~\cite{black202susie}}  & 87.0 & 69.0 & 49.0 & 38.0 & 26.0 & 2.69 \\
    \multicolumn{2}{l|}{GR-1~\cite{wu2023gr1}} & 85.4 & 71.2 & 59.6 & 49.7 & 40.1 & 3.06 \\
    \multicolumn{2}{l|}{OpenVLA$^{*}$~\cite{kim2024openvla}} & 91.3 & 77.8	 & 62.0	&52.1	&43.5 & 3.27 \\
    \multicolumn{2}{l|}{CLOVER~\cite{bu2024clover}}  & 96.0 & 83.5 & 70.8 & 57.5 & 45.4 & 3.53 \\
    \multicolumn{2}{l|}{RoboDual~\cite{bu2024robodual}}  & 94.4 & 82.7 & 72.1 & 62.4 & 54.4 & 3.66 \\
    \midrule
    \multicolumn{2}{l|}{\baseline{\modelname (Ours)}} & \baseline{\textbf{95.5}} & \baseline{\textbf{85.8}} & \baseline{\textbf{75.4}} & \baseline{\textbf{66.9}} & \baseline{\textbf{56.5}} & \baseline{\textbf{3.80}}\\
    \bottomrule
    \end{tabular}
    }
\end{table}

During training, we jointly optimize all components of our generalist policy, encompassing the visual encoders, the large language model (LLM) backbone, and the token prediction head. We utilize a batch size of 1,024 (with a per-device batch size of 32) and maintain a constant learning rate of $2e-5$. Empirical results indicate that 20,000 optimization steps are sufficient to achieve robust downstream performance, requiring approximately 30 hours of computation on a cluster equipped with 32 NVIDIA A100 GPUs. For pretraining on the ``Human'' and ``Bridge'' datasets (as presented in Table~\Cref{tab:libero}), we employ a global batch size of 258 distributed across 8 GPUs, totaling approximately 200 A100 GPU-hours.

\begin{figure*}[t]
    \centering
    \includegraphics[width=0.75\linewidth]{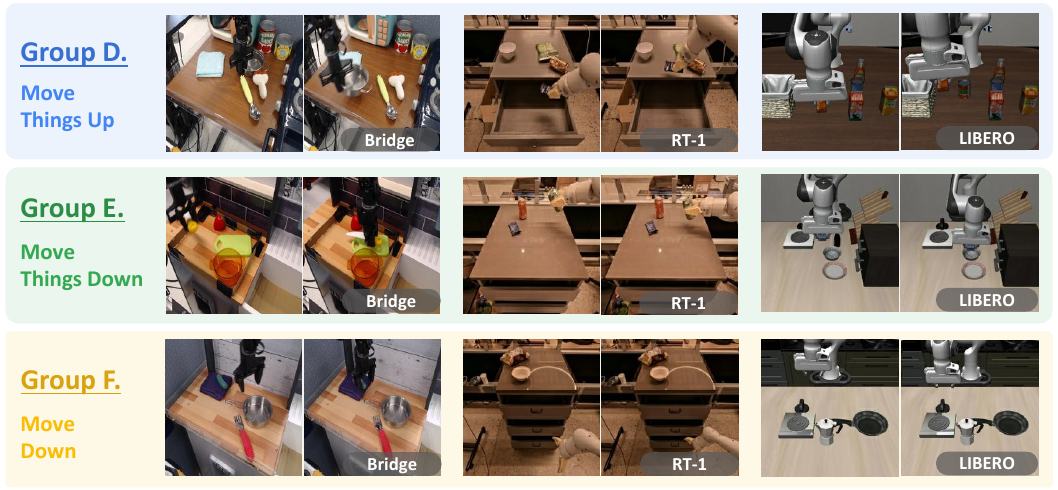}
    \caption{\textbf{Latent action analysis.} We show more image pairs labeled the same latent action from different source of data and embodiments. Each group of latent action presents semantic-consistent action}
    \label{fig:latent_action_2}
\end{figure*}

\begin{table*}
  [t]
  \centering
  \caption{\textbf{Grasp and task success rates on SimplerEnv}. As BridgeData is incorporated in our pretraining dataset, we investigate only training the decoder for adaptation (Decoder-only). UniVLA outperforms all baselines in success rate.}
  \resizebox{2\columnwidth}{!}{
  \begin{tabular}{l|cccccccccc}
    \toprule \multirow{2}{*}{Model}               & \multicolumn{2}{c}{\textbf{Put Spoon on Towel}}     & \multicolumn{2}{c}{\textbf{Put Carrot on Plate}} & \multicolumn{2}{c}{\textbf{Stack Green Block on Yellow Block}} & \multicolumn{2}{c}{\textbf{Put Eggplant in Yellow Basket}} & \textbf{\#Overall}                                         \\
                                                  & \begin{tabular}[c]{@{}c@{}}Grasp Spoon\end{tabular} & \begin{tabular}[c]{@{}c@{}}Success\end{tabular}  & \begin{tabular}[c]{@{}c@{}}Grasp Carrot\end{tabular}           & \begin{tabular}[c]{@{}c@{}}Success\end{tabular}            & \begin{tabular}[c]{@{}c@{}}Grasp Green Block\end{tabular} & \begin{tabular}[c]{@{}c@{}}Success\end{tabular} & \begin{tabular}[c]{@{}c@{}}Grasp Eggplant\end{tabular} & \begin{tabular}[c]{@{}c@{}}Success\end{tabular} & Average         \\
    \cmidrule{1-10} RT-1-X~\cite{brohan2022rt1}       & 16.7\%                                              & 0\%                                              & 20.8\%                                                         & 4.2\%                                                      & 8.3\%                                                     & 0\%                                             & 0.0\%                                                  & 0\%                                             & 1.1\%           \\
    Octo-Base~\cite{team2024octo}                 & 34.7\%                                              & 12.5\%                                           & 52.8\%                                                         & 8.3\%                                                      & 31.9\%                                                    & 0\%                                             & 66.7\%                                                 & 43.1\%                                          & 16.0\%          \\
    Octo-Small~\cite{team2024octo}                & 77.8\%                                              & 47.2\%                                           & 27.8\%                                                         & 9.7\%                                                      & 40.3\%                                                    & 4.2\%                                           & 87.5\%                                                 & 56.9\%                                          & 30.0\%          \\
    OpenVLA~\cite{kim2024openvla}                 & 4.1\%                                               & 0\%                                              & 33.3\%                                                         & 0\%                                                        & 12.5\%                                                    & 0\%                                             & 8.3\%                                                  & 4.1\%                                           & 1.0\%           \\
    RoboVLM~\cite{li2024robovlm} & 54.2\%                                              & 29.2\%                                           & 25.0\%                                                         & 25.0\%                                                     & 45.8\%                                                    & \textbf{12.5\%}                                          & 58.3\%                                                 & 58.3\%                                          & 31.3\%          \\
        \rowcolor[HTML]{EFEFEF} UniVLA (Decoder-only)   & 70.8\%                                              & 37.5\%                                           & 62.5\%                                                         & 33.3\%                                                     & 37.5\%                                                    & 4.2\%                                          & 87.5\%                                                & 66.7\%                                         & 35.4\% \\
    \rowcolor[HTML]{EFEFEF} UniVLA   & \textbf{76.4\%}                                              & \textbf{52.8\%}                                           & \textbf{79.2\%}                                                         & \textbf{55.6\%}                                                     & \textbf{66.7\%}                                                    & 2.8\%                                          & \textbf{93.0\%}                                                & \textbf{80.6\%}                                         & \textbf{47.9\%} \\
    \bottomrule
  \end{tabular}
  } 
  \label{tab:simplerenv_windowx}
  \vspace{-3ex}
\end{table*}

\subsection{Additional Results}
\label{sec:additional_results}
\subsubsection{CALVIN}

\paragraph{Experiment setup.} CALVIN~\citep{mees2022calvin} encompasses 34 distinct tasks, characterized by unconstrained task instructions that span a spectrum of skills, ranging from basic pick-and-place operations to articulated object manipulation. The benchmark includes four distinct environments, each featuring a Franka Panda robotic arm for tabletop manipulation tasks. In our study, we adopt the challenging evaluation setting, wherein policies are trained using demonstrations from environments A, B, and C, followed by zero-shot evaluations in environment D. The evaluation protocol comprises a test set of 1,000 unique instruction chains, each consisting of five consecutive tasks, designed to rigorously assess the generalization capabilities of the policies.

For OpenVLA, we finetune the officially provided checkpoint with LoRA~\cite{hu2021lora} for 200k steps and use an action chunk with size 8 to maximize performance. \modelname is optimized for 100k steps with a batch size of 128. We use a learning rate of $1.5e-4$ for the first 80k steps, and $1.5e-5$ for the rest. Similar to LIBERO experiments, we only take as inputs third-view RGB images and language instructions.

\paragraph{Results.} The results in~\Cref{tab:calvin} demonstrate UniVLA’s state-of-the-art performance in language-conditioned visuomotor control. UniVLA achieves 56.5\% success rate for completing all five tasks in sequence, surpassing the prior best method, CLOVER (45.4\%) by 11.1\%, and OpenVLA by 13\%. The average number of consecutively completed tasks increases from OpenVLA's 3.27 to 3.80. Notably, UniVLA’s performance gap widens progressively with task length, reflecting its ability to tackle complex, long-horizon manipulation tasks.

\begin{figure}[t]
    \centering
    \includegraphics[width=0.92\linewidth]{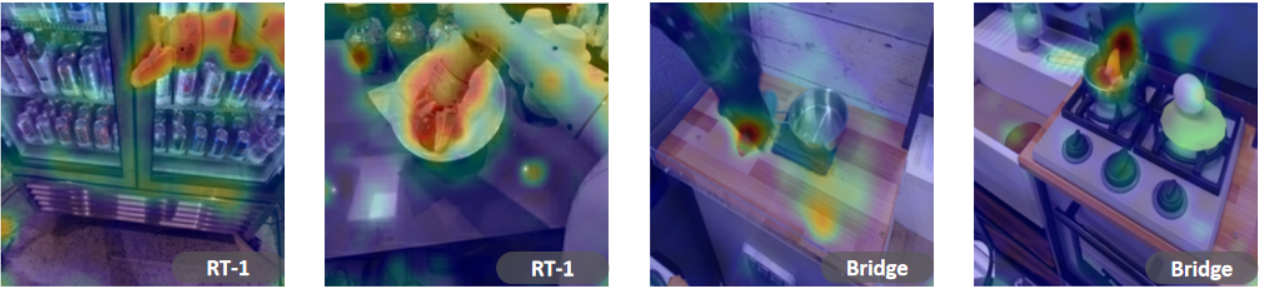}
    \caption{\textbf{Task-centric Latent action analysis.} We show the attention heatmap between task-centric latent actions and image patches, demonstrating concentrated focus on the robotic end-effector and target objects.}
    \vspace{-12pt}
    \label{fig:task-centric_analysis}
\end{figure}

\vspace{0.5em}
\subsubsection{SimplerEnv}

\paragraph{Experiment setup.} SimplerEnv~\cite{li2024simplerenv} are developed to genuinely reflect the performance of real-world policies by mirroring physical dynamics and visual appearances. We focus on four tasks concerning the ``WidowX + Bridge'' setup: 1) ``Put spoon on table cloth'', 2) ``Put carrot on plate'', 3) ``Stack green cude on the yellow cude'', and 4) ``Put eggplant in basket''. The pose and position of objects to be grasped will be randomly initialized given different seeds. Given that Bridge-V2~\cite{walke2023bridgedata} data is included in our pretraining dataset, we investigate training the action decoder head exclusively while keeping the remaining components of our model fixed, denoted as \textit{decoder-only} in~\Cref{fig:bridge-result}. Additionally, we perform further fine-tuning using LoRA~\cite{hu2021lora} on the complete Bridge-V2 dataset and evaluate the resulting policy. Our evaluation follows the pipeline proposed by~\citet{li2024robovlm}, wherein each task is assessed over 24 independent trials to ensure robust performance metrics.

\paragraph{Results.} The results in~\Cref{tab:simplerenv_windowx} underscores UniVLA’s superior performance in both grasp and task success rates on the SimplerEnv-Bridge benchmark, even under the constrained decoder-only adaptation setting. Specifically, decoder-only adaptation achieves a 35.4\% success rate, demonstrating its ability to retain pretrained knowledge while minimizing adaptation costs. However, full fine-tuning results in a reduced grasp rate compared to decoder-only training, likely due to overfitting to seen scene layouts in training samples. Overall, \modelname achieves a 42.7\% task success rate, outperforming OpenVLA and Octo-Base by 41.7\% and 26.7\%, respectively.

\begin{table*}[tb!]
 \centering
  \caption{
  \textbf{Scores of tasks.} Each sub-goal corresponds to one point.
  }
  \label{tab:task_score}
  \scalebox{0.81}{
\begin{tabular}{ccc}
    \toprule
    \textbf{Task Name} & \textbf{Total Score} & \textbf{Sub-goals} \\
    \midrule
    \multirow{3}{*}{Store the screwdriver} & \multirow{3}{*}{3} & Pick up the screwdriver. \\
     & & Place the screwdriver into the cabinet. \\
     & & Close the cabinet \\
     \midrule
    \multirow{3}{*}{Clean the cutting board} & \multirow{3}{*}{3} & Pick up the broom. \\ 
     & & Sweep the items into dustpan. \\
     & & Sweep \textbf{all} items into dustpan. \\
     \midrule
    \multirow{3}{*}{Fold towel twice} & \multirow{3}{*}{3} & Grasp the correct edge of the towel. \\
     & & Fold towel for the first time. \\
     & & Fold towel for the second time. \\
     \midrule
    \multirow{3}{*}{Stack tower of hanoi} & \multirow{3}{*}{3} & Choose the right tower. \\
     & & Stack the medium tower on top of the large one. \\
     & & Stack the small tower on top of the medium one. \\
    \bottomrule
\end{tabular}}
\end{table*}

\begin{table*}[t!]
    \centering
    \caption{\textbf{Experiment result.} We \textbf{bold} the best result and \underline{underline} the second. }
    \label{tab:detailed_exp}
    \small
    \setlength{\tabcolsep}{3.5mm}
    \scalebox{0.91}{
    \begin{tabular}{l|cccccccc|cc}
    \toprule
          \multirow{2}{*}{Method} &  \multicolumn{2}{c}{\makecell[c]{Store screwdriver}} & \multicolumn{2}{c}{\makecell[c]{Clean cutting board}} & \multicolumn{2}{c}{\makecell[c]{Fold towel twice}} & \multicolumn{2}{c|}{\makecell[c]{Stack tower of hanoi}} & \multicolumn{2}{c}{Average $\uparrow$}\\
         &Succ. & Score &Succ. & Score &Succ. & Score &Succ. & Score &Succ.  & Score \\
    \midrule
    Diffusion Policy~\cite{chi2023diffusion} & 40.0 & 1.20 & 33.3 & 0.67 & \textbf{53.3} & \underline{2.33} & 6.7 & 1.6 & 33.3 & 1.45\\
    OpenVLA~\cite{kim2024openvla} & 40.0 & 1.47 & 53.3 & 1.27 &33.3 & 1.87 &26.7 & 1.93 & 38.3 & 1.63\\
    LAPA (OXE)~\cite{ye2024lapa} & 60 & 2.0 & 40 & 1.47 & 33.3 & 2.2 & 46.7 & 2.13 & 45 & 1.95 \\
    \midrule
     \modelname (Bridge) & 66.7 & 2.13 & \underline{73.3} & 1.87 & 33.3 & 2.07 &46.7 & 2.13 & 55.0 & 2.05\\
      \modelname (OXE) & \underline{80.0} & \underline{2.73} & \underline{73.3} & \underline{1.93} & 33.3 & 2.13 & \underline{60.0} & \underline{2.60} & \underline{63.3} & \underline{2.35}\\
     \baseline{\modelname (Full)}    & \baseline{\textbf{93.3}} & \baseline{\textbf{2.87}} & \baseline{\textbf{100.0}} & \baseline{\textbf{2.33}} & 
     \baseline{\underline{46.7}} & \baseline{\textbf{2.47}} & \baseline{\textbf{86.7}} & \baseline{\textbf{2.87}} & \baseline{\textbf{81.7}} & \baseline{\textbf{2.63}} \\

    \bottomrule
    \end{tabular}}
\end{table*}

\vspace{0.5em}
\subsubsection{More Visualization}
\label{app:vis}

\paragraph{Additional examples for latent action analysis}.
As discussed in~\Cref{sec:navi}, We explore the cross-domain transferability of latent actions by presenting image pairs from diverse data sources that share the same latent action. Additional examples with distinct actions are provided in \Cref{fig:latent_action_2}.

\paragraph{Task-centric Latent action analysis.} We visualize the attention maps between learned task-centric latent actions and image patches in~\Cref{fig:task-centric_analysis}. The heatmaps reveal concentrated attention on task-critical regions: the robotic arm’s end-effector (e.g., gripper) and interacted objects (e.g., egg), while ignoring irrelevant background. This demonstrates that the latent action inherently encodes task-centric spatial priors, focusing only on entities necessary for downstream learning.

\subsection{Real-world Robots}
\label{sec:details_real_robots}

\subsubsection{Task setup and evaluation}

In task ``Store the screwdriver'', we randomly placed the screwdriver in three different positions for position generalization in training data, and tested it at four positions during evaluation. In task ``Store the screwdriver'', we randomly placed items on the cutting board, some of which in some cases could be swept into the dustpan in a single motion, while others required two sweeps during data collection. For task ``Fold towel twice'', we use a 20cm $\times$ 20cm towel and lay it flat on the table. During both training and testing, we randomly rotated the towel by different angles for evaluating generalization. In ``Stack tower of hanoi'', we randomly shuffle three cups and cover six different arrangements during both training and testing.

\begin{table}[tb!]
    \centering
    \caption{\textbf{Architecture details of action decoder in real-world experiment.} Additional proprioceptive state projection module is only adopted in real-world experiments.}
    \label{tab:architecture_decoder}
    \small
    \begin{tabular}{l|l|c}
    \toprule
    \multicolumn{3}{c}{Architecture of Action Decoder} \\
    \midrule
         \multirow{4}{*}{\makecell[l]{Latent Action\\ Attention Pooling}} 
         &Heads& 8\\
         & Head Dim. & 64 \\
         &Hidden Size & 512\\
         & MLP Ratio& 4 \\
         \midrule
         \multirow{4}{*}{\makecell[l]{Visual Embedding\\ Attention Pooling}} 
         &Heads& 8\\
         & Head Dim. & 64 \\
         &Hidden Size & 512\\
         & MLP Ratio& 4 \\
         \midrule
                  \multirow{2}{*}{\makecell[l]Action Projection} & Layers & 1 \\
         &Hidden Size & 512\\

         \midrule
         \multirow{2}{*}{\makecell[l]Proprio. Projection} & Layers & 2 \\
         &Hidden Size & 512 \\

          \midrule
         \multicolumn{2}{c|}{Parameters} & 12.6M\\
    \bottomrule
    \end{tabular}
\end{table}

We evaluate policies using a combined metric of success rate and task-specific scoring. \Cref{tab:task_score} shows the detailed scoring criteria. Full experimental results are presented in \Cref{tab:detailed_exp}. Notably, certain policies achieve identical success rates but exhibit divergent scores, reflecting differences in their performance quality. 

Single-task-trained methods such as Diffusion Policy perform well in tasks like Fold towel twice, which demand smooth, continuous, and highly structured action sequences. Although Diffusion Policy achieves a higher success rate in this task, its score remains lower than \modelname due to limited generalization across varying rotation angles. In contrast, \modelname’s generalizability enables robust performance across diverse positional configurations, resulting in a higher score even in cases of partial task completion.

\subsubsection{Architecture of the action decoder}

In the design of the action decoder architecture for both LAPA and our method, we use 2 multi-head attention blocks to process the latent action and visual embeddings~(also refer to~\Cref{sec:step3}), with a MLP layer to process proprioceptive states. The resulting embeddings are then concatenated and mapped to the desired action dimensions through a projection layer. Detailed parameters are shown in \Cref{tab:architecture_decoder}.

\end{document}